\documentclass[acmtog,nonacm]{acmart}

\usepackage{soul}
\usepackage{multirow}
\usepackage[inline]{enumitem}
\usepackage{subcaption}
\usepackage{float}

\author{Daniel Franzen}
\affiliation{%
  \institution{Johannes Gutenberg University Mainz}
  \city{Mainz}
  \country{Germany}}
\email{dfranz@uni-mainz.de}

\author{Jean Philip Filling}
\affiliation{%
  \institution{Johannes Gutenberg University Mainz}
  \city{Mainz}
  \country{Germany}}
\email{jefillin@uni-mainz.de}

\author{Michael Wand}
\affiliation{%
  \institution{Johannes Gutenberg University Mainz}
  \city{Mainz}
  \country{Germany}}
\email{Michael.Wand@uni-mainz.de}

\title{Discretizing Group-Convolutional Neural Networks for 3D Geometry in Feature Space}

\begin{document}

\begin{teaserfigure}
    \centering
    \includegraphics[width=1\linewidth]{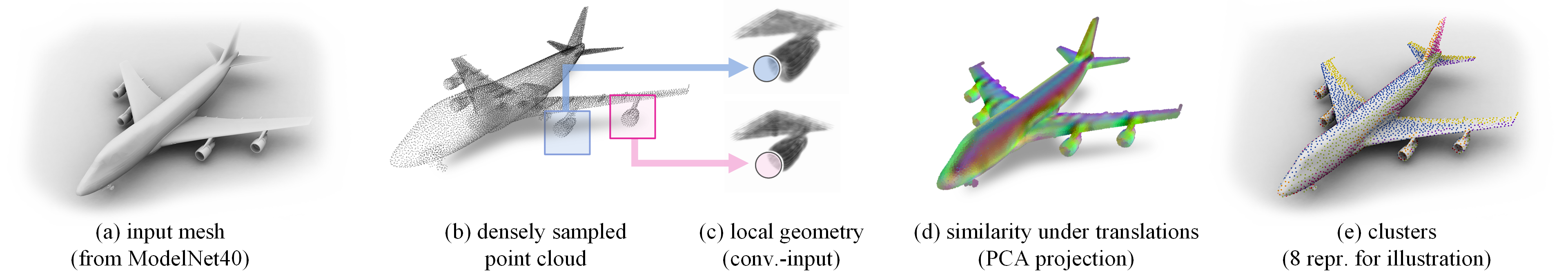}
    \caption{Typical 3D shapes (a) have a lot of locally (approximately) similar geometry (b). We detect similar regions (c,d) and compress the computations of Group-Convolutional Networks (GCNNs) in feature space, reusing computations on similar geometry (e). This significantly reduces costs of the linear representations that (G-)CNNs use for equivariance, in particular for larger groups such as $SE(3)$.}
    \label{fig:teaser}
\end{teaserfigure}

\begin{abstract}
Group-convolutional neural networks (GCNNs) are among the most important methods for introducing symmetry as an inductive bias in deep learning: In each linear layer, GCNNs sample a transformation group $G$ densely and correlate data and filters in different poses (with suitable anti-aliasing for steerable GCNNs) to maintain equivariance with respect to $G$. Unfortunately, applying filters to many data items resulting from this sampling is expensive (even for translations alone, i.e., in ordinary CNNs), and costs grow exponentially with increasing degrees of freedom (such as translations and rotations in 3D), which often hinders practical applications. In this paper, we propose sampling in feature space, i.e., replacing geometrically dense samples with representative samples selected by feature similarity. This decouples geometric resolution from memory and processing costs during training and inference, providing a novel way to trade off computational effort and accuracy. Our main empirical finding is that a coarse feature-space sampling already preserves classification accuracy remarkably well, which permits precomputation based on geometric similarity, accelerating the training of equivariant 3D classifiers substantially.
\end{abstract}

\maketitle

\let\oldthefootnote\thefootnote
\renewcommand{\thefootnote}{}
\footnotetext{Preprint. Under review.}
\let\thefootnote\oldthefootnote

\section{Introduction}

Deep learning \cite{Goodfellow-et-al-2016} has become an indispensable tool for shape understanding. Despite significant advances, deep learning with geometry still remains more challenging than other modalities such as image or text. Aside from data availability, the cost of representing and processing 3D shapes remains a central problem~\cite{thengane2025foundational}.
Our paper addresses the question of more efficient geometric representations, and we consider a specific problem: Reducing the costs of linear group representation layers.

Symmetry (invariance) fundamentally underlies all statistical learning. Generalization hinges upon finding similar statistics in novel conditions that are connected to training by an invariance of the probabilistic model. In addition to assuming the same, i.i.d.\ distribution of data items, most models introduce additional symmetry as an explicit inductive bias. For example, CNNs assume symmetry under translations, and transformers break their general permutation invariance through positional encoding.

Methods for introducing symmetry into networks broadly fall into two categories: geometric invariants and (linear) representations of the symmetry group. The first approach is to only ever input invariant (e.g. distances, angles) or covariant (e.g., geometrically constructed vectors or higher-order tensors) features into a network; a prominent example is graph neural networks~\cite{SchNet,Satorras21}. The alternative is to consider a single operation (linear layer) and apply the same operation repeatedly under all transformations of the symmetry group, yielding an \textit{equivariant} representation, i.e., input and output now behave homomorphically under group actions \cite{Cohen-GCNN-ICML-2016}. In the case of continuous Lie-groups, such as $SE(3)$, anti-aliasing is added (corresponding to subgroups represented by generalized Fourier-bases), which yields steerable group convolutional neural networks~\cite{Cohen-3DSteerable-2018}. Almost any practical implementation of symmetry in neural network architectures is based on one or a combination of both of these principles.

Linear representation layers come with serious costs: Sampling transformations densely is expensive, and costs grow exponentially with the dimension of the Lie algebra. Even with common sparsity optimizations (translations only along surfaces, exploit normal vectors to pre-align rotations), serious scaling issues remain. 
The objective of our paper is to reduce these costs. We propose and examine a very simple algorithmic building block that could in principle be plugged into any such scenario: Rather than discretizing the space of transformations at a uniform sampling rate, we discretize the space of input features by employing clustering. Afterwards, we aggregate computational paths in the network that operate on similar inputs and approximate the original computation at a fraction of the costs that is as such independent of the complexity of the symmetry group.

Two conditions need to be met to gain efficiency: We need to be able to compute similarity efficiently, and there must be redundancy in the input features. For the first part, we use a simple and generic ``standard''-strategy: clustering of high-dimensional vectors based on approximate nearest neighbor search~\cite{Frahm2010}. The second condition is not generic, but application dependent. In our paper, we explore, as a proof-of-concept, frame-invariant point cloud classifiers based on (group-) convolutional point cloud networks \cite{Atzmon-PointConv-Siggraph-2018}. We deliberately consider simple, canonical architectures (with pooling and batch-norm, but no further optimizations) to isolate the effect of clustering. Here, when running solely geometry-based clustering in preprocessing, we obtain speedups of $2$--$3\times$ for translations and roughly $10\times$ for $E(3)$-invariant networks (overall time, including preprocessing, at similar accuracy). Per-epoch speedups reach $30\times$--$45\times$ in the latter case, processing up to $256\times$ fewer features, reducing the burden on GPU memory. These results are observed for both synthetic meshes (ModelNet40) and real-world point clouds (ScanObjectNN, with and without color information). We also find clear limitations of feature-space discretization. In architectures that process high-resolution samples with large receptive fields, such as output layers of U-Nets, compression is not effective due to lack of redundancy.

In summary, our main contribution is a novel algorithmic building block for computing linear group representations in neural networks, with appealing memory and compute savings in feature-redundant scenarios such as shape classification.

\section{Related Work}
Considerable work has already been done on (more efficient) geometric deep learning:

\textbf{Geometric deep learning:} Soon after the breakthroughs in 2D image recognition~\cite{Krizhevsky2012}, applications on 3D geometry were proposed, starting with 3D CNNs on voxelized models~\cite{Song2014}. Insufficient resolution has been identified as a key factor in accuracy losses~\cite{QiCVPR2016}. Non-convolutional approaches build shift-invariant spatial fields relative to surface points~\cite{PointNet,PointNetPP}, processed by permutation-invariant arithmetic. Dynamic Graph CNN~\cite{dgcnn} extends this with dynamic $k$-NN graphs in feature space, capturing non-local relationships between semantically similar points. This is conceptually related to our compression, which also exploits feature-space similarity between spatially distinct regions; the mechanisms, however, are opposite. DGCNN uses feature space to determine \emph{connectivity} for message passing — with all points retained — and spends additional compute to do so, while our method uses it to determine \emph{partitioning} into shared representatives, reducing compute. Returning to CNNs, direct point convolution~\cite{Atzmon-PointConv-Siggraph-2018,PointCNN} has been introduced to reduce costs for 3D CNNs by a polynomial degree by restricting computation and activation to surfaces. For more expressivity, transformer architectures~\cite{Guo2021PCT,Zhao21PT} have been studied, where a positional encoding (based on invariants or linear representations) is used to encode geometric relationships. Self-supervised learning~\cite{Pang2022MAE} is a further direction, addressing the shortage of labeled training data.

We use a simple direct point-convolution network on surfaces, closely following \citeN{Wiersma-SurfaceCNNs-Siggraph-2020} and \citeN{kuipers2023se3}. This allows us to isolate the effect of the proposed feature-space compression scheme and study trade-offs without involving complex architectural choices. Our architecture does provide reasonable, middle-of-the-road performance, but is not close to top SOTA accuracy figures. Our key result is that the networks \textit{maintain} their original performance at a fraction of the training cost when compression is enabled.

\textbf{Rotationally-equivariant networks} have been considered in the seminal paper by \citeN{Cohen-GCNN-ICML-2016}, which introduces the framework of linear representation theory, with later extensions to steerable representations~\cite{Cohen-Steerable-2017,weiler2021coordinate}, including a construction with full $SE(3)$-equi\-vari\-ance~\cite{Cohen-3DSteerable-2018}. A large number of follow-up papers have refined this approach, see e.g. \citeN{Fei24Survey} for a survey. In parallel, \emph{regular} G-CNNs were developed, where feature maps are defined on a discrete sampling of $G$ and kernels are parameterized in the spatial domain using B-splines~\cite{bekkers2020bspline}, RBFs~\cite{kuipers2023se3}, or learned via continuous parameterizations~\cite{knigge2022separable}. These methods extend to point clouds via weight-sharing in position-orientation space~\cite{bekkers2024ponita}, to meshes via gauge-equivariant convolutions~\cite{dehaan2021gauge}, and to discrete subgroups via Platonic symmetry constructions~\cite{islam2025platonic,worrall2018cubenet,winkels2018gcnn3d}. Recent empirical studies compare regular and steerable variants directly~\cite{vadgama2025probing} and examine whether equivariance remains beneficial at scale~\cite{brehmer2024equivariance}. The regular variant can be understood as a change of basis within bandlimited functions, and we also follow this path to facilitate clustering.

\textbf{Graph networks:} Graph neural networks~\cite{SchNet,Satorras21,bronstein2021geometric} avoid the costs of linear group representations and use rigidly invariant geometric properties instead, such as distances, angles, scalar products, or, in more general terms, a geometric (Clifford) algebra~\cite{brehmer2023geometric,ruhe2023clifford}. Some architectures (popular in computational physics) employ hybrids of GNNs and local GCNNs based on linear representations with spherical harmonics and Clebsch-Gordon coefficients~\cite{Kondor18,TensorFieldNetworks,Fuchs20}.

\textbf{Efficiency:} Ballooning costs become already prominent in translational 3D CNNs~\cite{QiCVPR2016}, which motivated sparse approaches, operating directly on surfaces~\cite{Atzmon-PointConv-Siggraph-2018,MeshCNN}. The situation deteriorates when including rotations. A popular trick is to only represent $SO(2)$ by aligning rotations with surface normals \cite{Wiersma-SurfaceCNNs-Siggraph-2020}; we rely on this approach, too, for most $E(3)$-invariant experiments, as it is simple and effective (we do study costs in full $SO(3)$-convolutions briefly).
Steerable GCNNs represent features in the frequency domain, which requires a fast Fourier transform (FFT). Rotations around the normal and translations are handled by standard algorithms; for more complex group structures, such as $SO(3)$, generalized FFT algorithms~\cite{Vollrath2010,chen2021equivariant,Lee22} are known, but they are complex and software libraries remain limited. Alternatively, one can exploit separability across subgroup and spatial dimensions~\cite{knigge2022separable,kuipers2023se3}, the spatial sparsity of 3D data~\cite{lin2021sparse}, or operate directly on the steerable basis~\cite{cesa2022a}. All of this does not reduce the memory costs and still implies the same, higher-order polynomial growth in runtime with resolution. One can of course deliberately limit angular resolution, such as in E2PN~\cite{zhu2023e2pn}. A more directly related alternative is adaptive down-sampling, which aims at maintaining discriminative points to boost speed and accuracy \cite{Nezhadarya_2020_CVPR,bytyqi2020,metasampler,Wang21GDS,Li25}; these methods all target translational CNNs only, aiming at adaptive coarse-graining, but do not provide a general scheme for discretizing linear group representations.

\textbf{Attention:} The quadratic complexity of attention layers in transformers~\cite{Vaswani2017} is a major computational bottleneck. Correspondingly, merging and clustering of tokens has been one important avenue of reducing costs, often with minimal loss or even gains in performance~\cite{DeepSeek25}. General token-level merging~\cite{Wang21GDS,bolya2023token} and pruning \cite{rao2021dynamicvit} inside a transformer, beyond attention, has also been applied to 2D vision transformers to speed up computation. While the token-merging idea as such is related, these methods operate in token space, specific to transformers, and symmetry is captured only indirectly via positional embedding. Recent work explores group-equivariant transformers for 3D point clouds via discrete subgroups~\cite{islam2025platonic} and unified comparisons of regular and steerable architectures~\cite{vadgama2025probing}, but to our knowledge, no prior work targets compression of general steerable filters across continuous Lie groups in the way we do.

\textbf{(Approximate) partial symmetry detection} has been studied extensively in literature, and efficient algorithms for this exact task are what we need for our preprocessing step. The idea was introduced in the seminal paper by Mitra et al.~\shortcite{Mitra2006}, with numerous follow-up methods that increase efficiency and/or robustness. This includes methods that can find all rigid symmetries in 3D city scans in minutes to hours on a single PC~\cite{Kerber2013}; see  \cite{Mitra2013Survey} for a comprehensive survey. A general, scalable, and easy-to-implement strategy is to vectorize local geometry and use high-dimensional approximate nearest neighbor search to perform pairwise clustering in near-linear time (in practice)~\cite{Frahm2010,Kerber2013}. Using this approach verbatim on network inputs already yields the substantial speedups seen in our experiments, though we acknowledge that even more optimizations are possible (such as clustering continuous symmetry separately \cite{Gelfand2004,Kalojanov2012}).

\section{Compressed GCNNs for 3D Point Clouds}
\label{sec:method}

\subsection{Point Cloud Convolutions}
\label{sec:pcconv}

We start by defining a convolution operator that operates directly on point clouds~\cite{Atzmon-PointConv-Siggraph-2018,KPConv}:
Let $\{\mathbf{p}_i\}$ denote a set of input points with associated feature
vectors $\phi_{i,c}$, where $c$ indexes the $C$ feature channels, and
let $\{\mathbf{p}'_j\}$ denote a set of output points at which we wish to
compute output features $\phi'_{j,c'}$ with $C'$ channels. A point
cloud convolution can then be written as
\begin{equation}
  \phi'_{j,c'} \;=\; \sum_{i,k,c} w_{k,c,c'}\, \phi_{i,c}\,
  \sigma\!\left(\mathbf{p}'_j + \mathbf{d}_k - \mathbf{p}_i\right),
  \label{eq:pcconv}
\end{equation}
where $\{\mathbf{d}_k\}$ are the kernel offsets at which the convolution is
evaluated, $w_{k,c,c'}$ are the trainable convolution weights, and
$\sigma(\cdot)$ is a localized kernel function that attains its
maximum at the origin and decays with distance. While a Gaussian is a
natural choice, we use a B-spline kernel~\cite{splineCNN} in practice
because its compact support enables sparse evaluation.

The geometry-dependent factor in Eq.~\ref{eq:pcconv}—comprising the
input and output point clouds, the kernel offsets, and the kernel
function $\sigma$—can be collected into a single tensor
\begin{equation}
  T_{ijk} \;=\; \sigma\!\left(\mathbf{p}'_j + \mathbf{d}_k - \mathbf{p}_i\right),
  \label{eq:interaction}
\end{equation}
which we refer to as the \emph{interaction tensor}: it encodes how
input features at $\mathbf{p}_i$ contribute to output features at $\mathbf{p}'_j$ for
each kernel offset $\mathbf{d}_k$, given the geometry and the choice of
convolution parameters. Although $T_{ijk}$ is conceptually convenient,
its size grows with the product of the input and output point counts
and is therefore prohibitive to store explicitly for realistic point
clouds.

\subsection{Feature-Cluster Compression}
\label{sec:compression}

Our compression scheme reduces the storage and computational cost of
Eq.~\ref{eq:pcconv} by clustering point features and replacing them
with cluster representatives. Let $\pi(i) \in \{1,\dots,M\}$ denote
the cluster assignment of input point $\mathbf{p}_i$ and
$\pi'(j) \in \{1,\dots,N\}$ the cluster assignment of output point
$\mathbf{p}'_j$. We write
$\mathcal{C}_m = \{\, i : \pi(i) = m \,\}$ for the set of input points
in cluster $m$ and analogously
$\mathcal{C}'_n = \{\, j : \pi'(j) = n \,\}$ for the output points in
cluster $n$. The representative feature of each cluster is the mean
of its members' features:
\begin{equation}
  \hat\phi_{m,c}   = \frac{1}{|\mathcal{C}_m|}
                     \sum_{i \in \mathcal{C}_m} \phi_{i,c}, \quad\ 
  \hat\phi'_{n,c'} = \frac{1}{|\mathcal{C}'_n|}
                     \sum_{j \in \mathcal{C}'_n} \phi'_{j,c'}.
\end{equation}
Substituting the input representatives $\hat\phi_{\pi(i),c}$ for the
per-point features in Eq.~\ref{eq:pcconv} and averaging over each
output cluster yields
\begin{equation}
  \hat\phi'_{n,c'} \;=\; \frac{1}{|\mathcal{C}'_n|}
  \sum_{j \in \mathcal{C}'_n} \sum_{i,k,c}
  w_{k,c,c'}\, \hat\phi_{\pi(i),c}\, T_{ijk}.
  \label{eq:compressed-naive}
\end{equation}
At this stage we have reduced the feature storage to $M$ input and
$N$ output representatives plus the integer assignment maps, but the
convolution itself still depends on the full geometry through
$T_{ijk}$.

The key step is to reorder the sums so that the per-point indices are
absorbed into a smaller geometric object. Splitting the sum over $i$
according to its cluster assignment and pulling the cluster-constant
representative outside gives
\begin{equation}
  \hat\phi'_{n,c'} \;=\; \sum_{m,k,c} w_{k,c,c'}\, \hat\phi_{m,c}
  {\left(\frac{1}{|\mathcal{C}'_n|}
    \sum_{j \in \mathcal{C}'_n} \sum_{i \in \mathcal{C}_m}
    T_{ijk}\right)}.%
  \label{eq:compressed-conv}
\end{equation}
The bracketed term, which we call the \emph{compressed interaction
tensor} $\hat T_{mnk}$, depends only on the geometry and the cluster
assignments and can be precomputed.
Whereas $T_{ijk}$ scales with the number of input and output points,
$\hat T_{mnk}$ scales with the number of clusters and is small
enough to materialize and store. Provided the cluster assignments
remain fixed, $\hat T_{mnk}$ is identical across training epochs;
in our pipeline it is computed once during the first epoch, written
to disk, and reused for all subsequent epochs. This avoids ever
reconstructing the full point cloud from its representatives at
training time.
The forward pass through a compressed convolution layer reduces to
two contractions. We first apply the compressed interaction tensor
to the input representatives,
\begin{equation}
  h_{n,k,c} \;=\; \sum_m \hat\phi_{m,c}\, \hat T_{mnk},
  \label{eq:step1}
\end{equation}
and then apply the trainable weights,
\begin{equation}
  \hat\phi'_{n,c'} \;=\; \sum_{k,c} w_{k,c,c'}\, h_{n,k,c}.
  \label{eq:step2}
\end{equation}
A single reshape between the two steps is all that is required to
expose them as standard dense matrix multiplications, which map
directly onto efficient GEMM kernels on modern accelerators.

\subsection{Group-Equivariant Extension}
\label{sec:equivariant}

Eq.~\ref{eq:pcconv} describes a translation-equivariant point
convolution. The construction extends to more general group
equivariance by replacing point coordinates with group elements, and replacing vector algebra with group operations~\cite{bekkers2020bspline,bekkers2024ponita}. Let $g_i$ and $g'_j$
denote the input and output group elements and $\delta_k$ the kernel
offsets, now also group elements. The convolution becomes
\begin{equation}
  \phi'_{j,c'} \;=\; \sum_{i,k,c} w_{k,c,c'}\, \phi_{i,c}\,
  \sigma\!\left(g'_j \cdot \delta_k \cdot g_i^{-1}\right).
  \label{eq:gconv}
\end{equation}
The kernel $\sigma$ is now defined on the group and should attain its
maximum at the identity, decaying as one moves away from it.

For an $\mathrm{SE}(3)$-equivariant network—combining translation and
rotation—we factor the relative group element
$g'_j \cdot \delta_k \cdot g_i^{-1}$ into its translational and
rotational components, take the translation distance $d_T$ and the
rotation angle $d_R$ separately, and combine them into a scalar
distance with a balance factor $\lambda$:
\begin{equation}
  d_{\mathrm{SE}(3)} \;=\; \sqrt{\,d_T^{\,2} + \lambda\, d_R^{\,2}\,}.
  \label{eq:se3dist}
\end{equation}
Applying the B-spline kernel to $d_{\mathrm{SE}(3)}$
yields a localized kernel on $\mathrm{SE}(3)$, and the compression
machinery of Section~\ref{sec:compression} carries over without
modification: the interaction tensor and its compressed counterpart
are constructed exactly as before, with $\sigma$ now evaluated on
group distances rather than Euclidean ones.

\subsection{Normalization and Activation}
\label{sec:bnrelu}

Because the representatives $\hat\phi'_{n,c'}$ live in the same
feature space as the per-point features they summarize, pointwise
operations such as normalization and nonlinearities can be applied
to them directly. Strictly speaking this is an approximation:
$\mathrm{ReLU}$ and batch normalization do not commute with cluster
averaging, so applying them to representatives is not equivalent to
applying them per-point and re-averaging. In practice, however,
treating the representatives as ordinary feature vectors works well
and preserves the compression throughout the network. We follow
each compressed convolution with batch normalization and a ReLU
nonlinearity applied directly to $\hat\phi'_{n,c'}$; the end-to-end
accuracy reported in our experiments serves as the empirical
justification for this choice.

\subsection{Clustering}
\label{sec:clustering}

The cluster assignments $\pi$ and $\pi'$ determine both the
compression ratio and the accuracy of the compressed convolution.
We consider two families of schemes: clustering on activations from
a partially trained network, and clustering purely on geometry
without using the trainable weights.
 
\paragraph{Activation-based clustering.}
Within a single convolution block there are four natural points at
which the activations can be inspected and clustered:
\begin{enumerate*}[label=(\roman*)]
  \item immediately after applying the compressed interaction
    tensor, before the learned weights, i.e.\ on $h_{n,k,c}$;
  \item directly after applying the learned weights, on
    $\hat\phi'_{n,c'}$;
  \item after normalization; and
  \item after the nonlinearity.
\end{enumerate*}
All four couple preprocessing to the trained weights, since they
require running activations through a partially trained network.
 
\paragraph{Geometry-only clustering.}
We additionally derive a scheme that depends only on the geometry
of the point cloud and the network architecture. For the first
convolution layer this is straightforward: its inputs are constant
across training, so clustering can be performed once on $h_{n,k,c}$.
For subsequent layers the activations themselves depend on the
previous layer's weights, but the cluster \emph{assignments} of the
previous layer are available from preprocessing. We therefore
replace the per-point feature vectors with one-hot encodings of
their cluster ids from the preceding layer and apply the same
clustering procedure to these synthetic features. This can be
viewed as an unsupervised hierarchical decomposition of the point
cloud, performed entirely at preprocessing time and independent of
the trainable parameters.
 
\paragraph{Default choice.}
We adopt geometry-only clustering as our default; the empirical
comparison between all five variants is reported in
Section~\ref{sec:where-to-cluster}.

\section{Implementation \& Experimental Setup}

To evaluate the effect of our feature-space compression on computational costs and accuracy, we implement two different architectures: a classical pooling-based network for classification, which we evaluate on the ModelNet40~\cite{Wu2015} and ScanObjectNN~\cite{Uy-ScanObjectNN-iccv19} datasets, which are the most prominent standard benchmarks for 3D object classification, as well as a U-Net based architecture for class segmentation, which we evaluate on the ScanNet20~\cite{dai2017scannet} dataset. For ScanObjectNN, we use the ``OBJ\_ONLY'' variant.

\paragraph{Preprocessing}
ModelNet40 consists of synthetic triangle meshes, which we convert into 3D point clouds via Poisson-disc sampling before further processing. ScanObjectNN contains real-world 3D scanner data and already comes as point clouds. For all data, we further pre-generate different downsampling levels with Poisson-disc sampling, each containing one fourth of the points of the previous level ($16384$, $4096$, $1024$, $256$, $64$, $16$). We scale-normalize all models to a unit bounding cube, centered at the origin. Normals are taken for the dataset directly, with a simple orientational alignment away from center of the point cloud.

\paragraph{Classification Architecture:} All of our classification experiments use the same 6-layer network with $[24,32,48,64,80,96]$ output feature channels $c'$ for each layer. The input layer evaluates the grid of spatial filters on the points themselves; additionally, we add the 3 color values as additional channels in some of our ScanObjectNN experiments. We do not use normal information as input (however, we use normal information for aligning the filters in our rotationally equivariant networks). We employ BatchNormalization~\cite{Batchnorm-Ioffe-ICML-2015} and ReLU after each convolutional layer. Downsampling is not performed as a separate operation, but by using the next (lower sampled) precomputed sampling level as output points for the convolution operation. We always apply downsampling in the first layer, and add additional downsampling steps in the final layers as required to always reach 16 spatial points after the last convolutional layer. After the final convolution layer, we apply global average pooling over spatial points and rotations and then use a LogSoftmax layer to generate the class logits.

\paragraph{Segmentation Architecture:} For our segmentation architecture, we use a classical U-Net architecture with 3 downsampling steps, and use two convolutions at each level in the encoder and decoder path, as well as two convolutions at the bottom level. As in the classification architecture, each convolutional layer is followed by BatchNormalization and ReLU. After the final layer, we use LogSoftmax to generate class logits for each point.

\paragraph{Convolution Kernels:} We use quadratic B-spline kernels on a grid (e.g., $3 \times 3 \times 3$). For the classification architecture, we chose an initial spacing $\Delta = 0.02$, doubled in each subsequent layer. For the segmentation architecture, we use an initial spacing of  $\Delta = 0.01$, doubled at each downsampling step. The kernel width is always chosen to match the spacing so that adjacent kernels overlap and form a smooth partition of unity.

\paragraph{Rotation:} For all rotationally equivariant networks, we perform training and testing on input shapes in random rotations (sampled uniformly from $SO(3)$) if not stated otherwise. For the purely translational GCNNs, we keep the data aligned as provided in the datasets. Exceptions are the ablation studies in Fig.~\ref{fig:ModelNetTrStability}, \ref{fig:ScanObjTrStability} and \ref{fig:ModelNetEqStability}.

\paragraph{Training parameters:}
Our default training procedure is training for 15 epochs, starting with an initial learning rate of $10^{-3}$ and decaying by a factor of 2 after the epochs $[6, 9, 12, 13, 14]$. We use the Adam optimizer in combination with a softmax cross-entropy loss. After the training is finished, we do another pass on the full training set to fit the learned parameters of the BatchNormalization, keeping the rest of the network fixed.
For the full $SO(3)$ convolutions, as well as the segmentation architecture, we expand the training to 30 epochs and also stretch the learning rate schedule, decaying by a factor of 2 after epochs $[12, 18, 24, 26, 28]$.
Since we evaluate our segmentation experiments by mean class IoU instead of accuracy (as is common for ScanNet20), we modify the loss to adapt to this by using a mixed loss (25\% softmax cross-entropy + 75\% dice loss).
 
\paragraph{Clustering algorithm.}
We use $k$-means with a maximum of $10$ iterations to compute the cluster assignments $\pi$ and $\pi'$, with cluster centers initialized as a random subset of the points. These choices are justified by the ablation in Section~\ref{sec:clustering-algorithm}.

\paragraph{Technical aspects:} We implement all operations in Python using \emph{PyTorch}. Performing direct point convolution involves sparse matrix algebra, which is not fully optimized in PyTorch. We thus employ \emph{pyKeOps} for the convolution operation on the point clouds, which compiles custom CUDA kernels for generic reductions on the fly, with support for block-sparsity. The $k$-means assignment step is computed on GPU using the \texttt{MaxSimCuda} kernel from \emph{TorchPQ} with Euclidean distance. We will release our code under an open-source license upon publication.

\paragraph{Technical scope:} We run all experiments on a single NVIDIA RTX 4090 GPU, in a system with an AMD Ryzen 16-core CPU and 128 GB RAM. We do not perform experiments that exceed the available GPU RAM of 24~GB during training (we do not attempt gradient checkpointing, multi-GPU setups etc. to fit larger networks). We also restrict the scope of the experiments to settings where training can be performed within a few hours at most. In this way, we can get a realistic impression of the benefits of our proposal under plausible resource constraints.

\newcommand{\V}{\text{Pts.}}

\begin{table*}[t]
\small
\centering
\caption{Overall test set accuracies and runtimes on ModelNet40 and ScanObjectNN for a translation-only equivariant model (i), our optimized SE(3) equivariant model (ii), and full $SO(3)$ convolutions (iii). The first training epoch includes preprocessing and is thus reported separately for compressed runs.}
\label{tab:results}
\begin{tabular}{cccrrrcrrr}
\toprule
\multirow{2}{*}{\centering Surfels} &
\multirow{2}{*}{\centering Repr.} &
\multicolumn{4}{c}{ModelNet40} & 
\multicolumn{4}{c}{ScanObjectNN (OBJ\_ONLY, not using colors)} \\
\cmidrule(lr){3-6} \cmidrule(lr){7-10}
 &  & Accuracy & Epoch time & 1st Epoch & Total & Accuracy & Epoch time & 1st Epoch & Total \\
\midrule

\multicolumn{10}{c}{(i) Translation-only equivariant model (trained 15 epochs)} \\
\midrule
$256\,\V$
& n.n.
& 86.39\% &   00:09 & &   02:15
& 73.24\% &   00:03 & &   00:45 \\

$1024\,\V$
& n.n.
& 89.87\% &   00:13 & &   03:15
& 84.73\% &   00:05 & &   01:15 \\

$4096\,\V$
& n.n.
& 90.15\% &   00:22 & &   05:45
& 86.79\% &   00:06 & &   01:30 \\

$16384\,\V$
& n.n.
& 91.21\% &   01:12 & &   18:00
& 87.65\% &   00:16 & &   04:00 \\

$4096\,\V$
& 16
& 88.37\% &   00:03 & 00:23 &   01:05
& 83.36\% &   00:01 & 00:06 &   00:26\\

$4096\,\V$
& 32
& 89.59\% & 00:04 & 00:32 &   01:28
& 85.08\% & 00:01 & 00:09 &   00:30 \\

$4096\,\V$
& 64
& 89.99\% & 00:06 & 00:56 &  02:20
& 85.76\% & 00:02 & 00:13 &  00:37  \\

$4096\,\V$
& 128
& 90.24\% & 00:10 & 02:01 &  04:21
& 86.62\% & 00:02 & 00:28 &  01:03  \\

$16384\,\V$
& 128
& 90.56\% & 00:10 & 04:40 &  07:00
& 87.48\% & 00:02 & 01:04 &  01:39 \\

\midrule
\multicolumn{10}{c}{(ii) Optimized SE(3) equivariant model (normal-aligned filters, trained 15 epochs)} \\
\midrule
512 ($64\,\V \times 8\,\text{Rot.}$) 
& n.n.
& 69.08\% &   03:11 & &   47:45
& 52.32\% &   00:42 & &   10:30 \\

2048 ($256\,\V \times 8\,\text{Rot.}$) 
& n.n.
& 83.75\% &   12:19 & & 3:04:45
& 78.04\% &   02:49 & &   42:15 \\

8096 ($1024\,\V \times 8\,\text{Rot.}$) 
& n.n.
& 86.35\% &   57:10 & & 14:17:30
& 84.05\% &   13:28 & &  3:22:00 \\

32768 ($4096\,\V \times 8\,\text{Rot.}$)
& n.n.
&         &         & &
& 85.59\% &   56:46 & & 14:11:30 \\

32768 ($4096\,\V \times 8\,\text{Rot.}$)
& 16
& 85.45\% &   00:22 &   59:51 & 1:04:59
& 82.68\% &   00:05 &   13:08 &   14:18 \\

32768 ($4096\,\V \times 8\,\text{Rot.}$)
& 32
& 85.78\% &   00:33 & 1:34:23 & 1:42:05
& 84.73\% &   00:12 &   20:43 &   23:31 \\

32768 ($4096\,\V \times 8\,\text{Rot.}$)
& 64
& 86.83\% &   01:16 & 2:55:06 & 3:12:50
& 84.91\% &   00:30 &   38:28 &   45:28 \\

32768 ($4096\,\V \times 8\,\text{Rot.}$)
& 128
& 87.07\% &   04:03 & 5:24:43 & 6:21:25
& 85.42\% &   01:30 & 1:11:25 & 1:32:25 \\

\midrule
\multicolumn{10}{c}{(iii) Full SO(3) group convolution (trained for 30 epochs)} \\
\midrule

11776 ($256\,\V \times 46\,SO(3)$)
& 64
& 65.84\% &   09:44 & 3:58:42 & 8:40:50
& 56.09\% &   02:17 &   57:23 & 2:03:36 \\

47104 ($1024\,\V \times 46\,SO(3)$)
& 64
& 68.12\% &   10:09 & 11:45:23 & 16:39:53
& 63.81\% &   02:23 &  2:45:47  &  3:54:54 \\

\bottomrule
\end{tabular}
\end{table*}

\section{Results}
\label{sec:results}
We evaluate our method on three tasks. Section~\ref{sec:results-classification-tr} reports classification results with a translation-equivariant architecture, and Section~\ref{sec:results-classification-eq} extends these to fully SO(3) equivariant architectures; both are evaluated on synthetic objects from ModelNet40 and real 3D scans from ScanObjectNN. Section~\ref{sec:segmentation-experiments} reports semantic segmentation on ScanNet20. We further analyze the accuracy–runtime trade-off of our compression and provide ablations on clustering choices in Section~\ref{sec:clustering-ablation}.

\subsection{Classification with translational equivariance}
\label{sec:results-classification-tr}
For the purely translational equivariant case, computational costs are still small, so they are no practical concern yet. However, our compression method can already provide moderate run-time benefits (see Table~\ref{tab:results} (i)). For the ScanObjectNN dataset with an input point cloud size of 16384 points, standard training for 15 epochs takes about 4 minutes. Using our compression method with 128 representatives, we can reduce overall training time by a factor of $\sim 2.4$, with only a negligible drop in accuracy (87.65\% vs 87.48\%). Per-epoch runtime, excluding preprocessing, drops by a factor of 8, with the number of features processed reduced by a factor of $128$.

\begin{figure}[H]
  \centering
  \begin{subfigure}[b]{0.33\linewidth}
    \includegraphics[width=\linewidth]{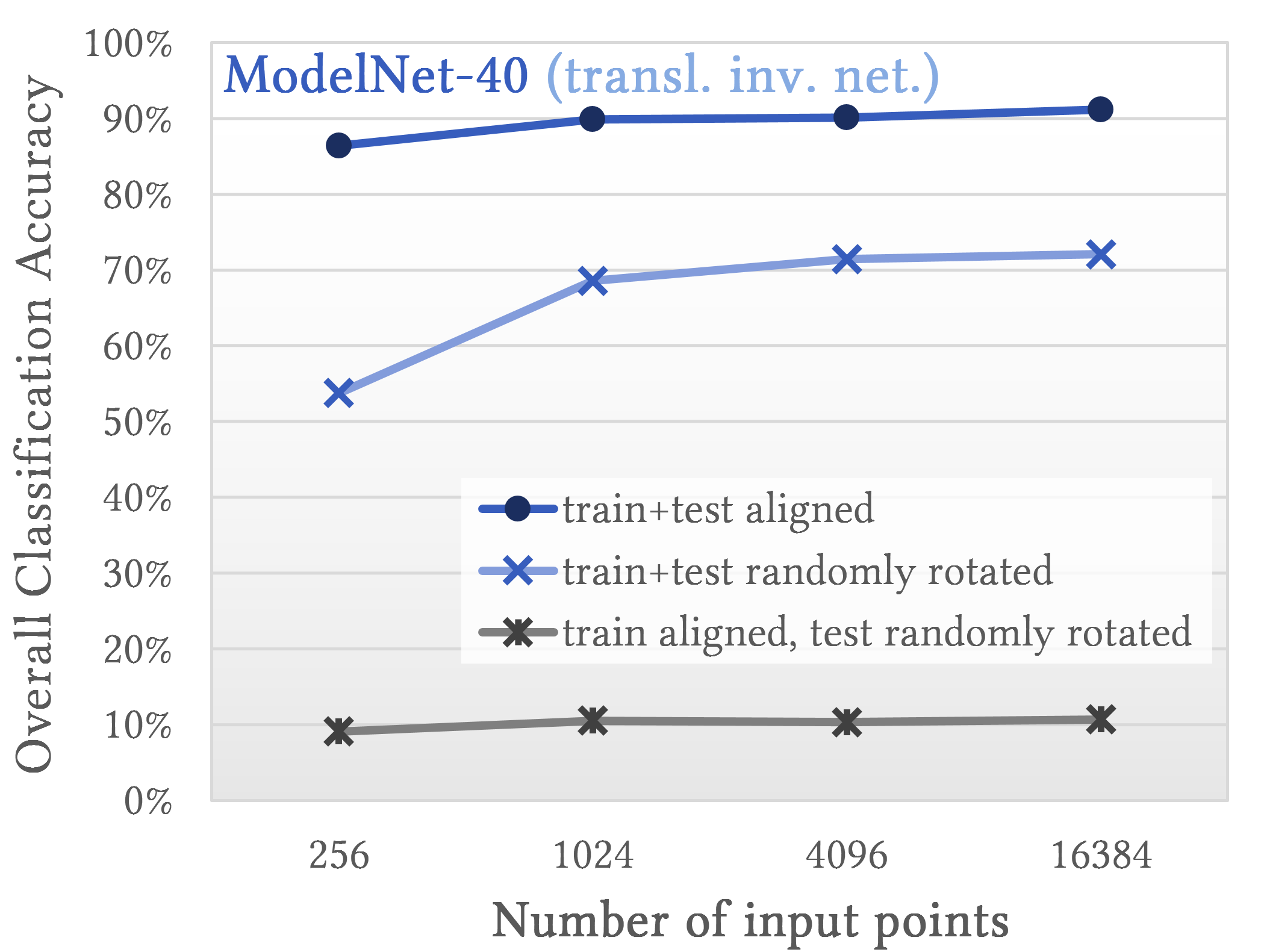}
    \caption{Uncompressed baseline.}
    \label{fig:ModelNetTrStability}
  \end{subfigure}
  \hfill
  \begin{subfigure}[b]{0.33\linewidth}
    \includegraphics[width=\linewidth]{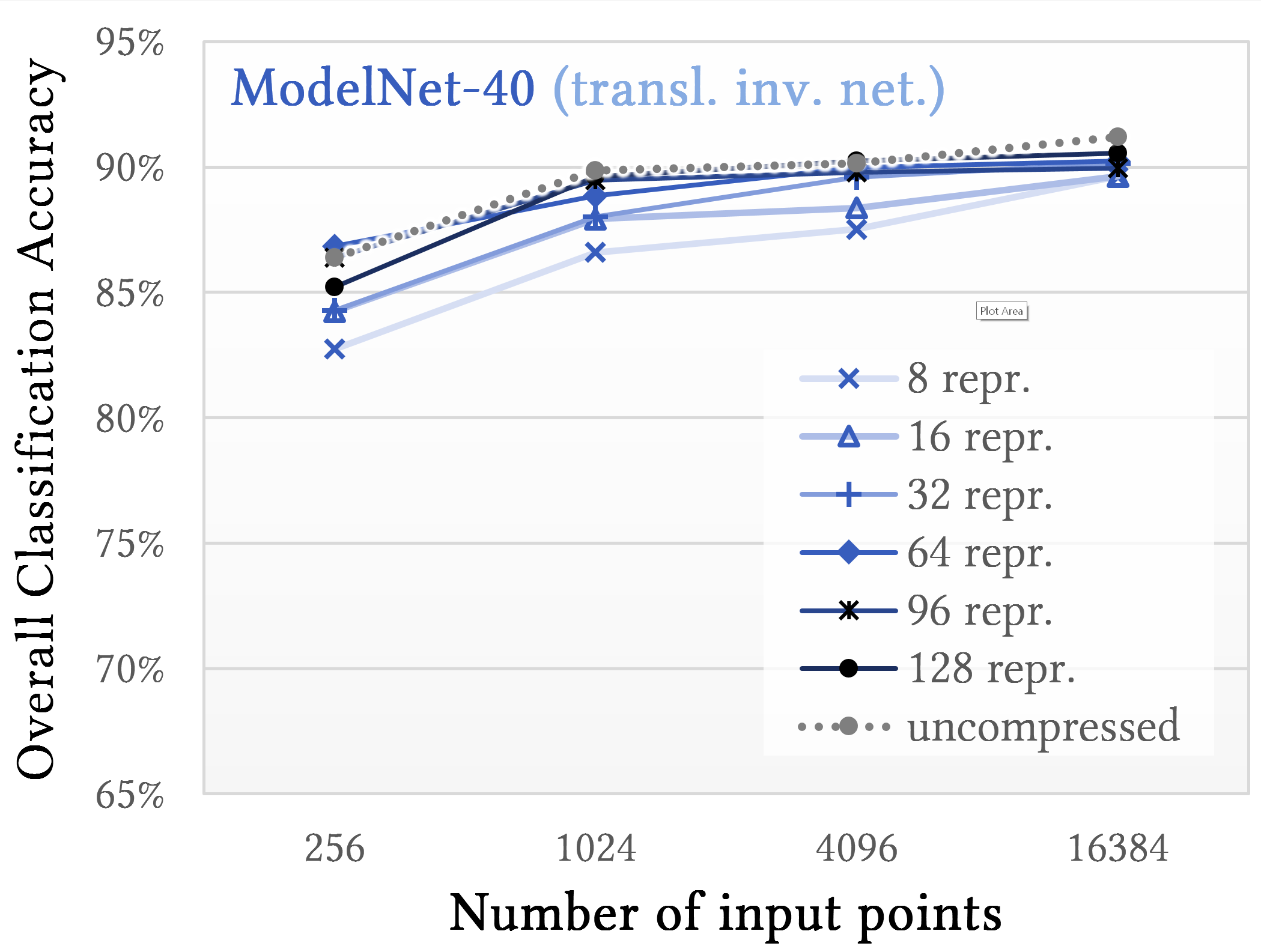}
    \caption{Compression accuracy loss.}
    \label{fig:ModelNetTrAccuracy}
  \end{subfigure}
  \hfill
  \begin{subfigure}[b]{0.33\linewidth}
    \includegraphics[width=\linewidth]{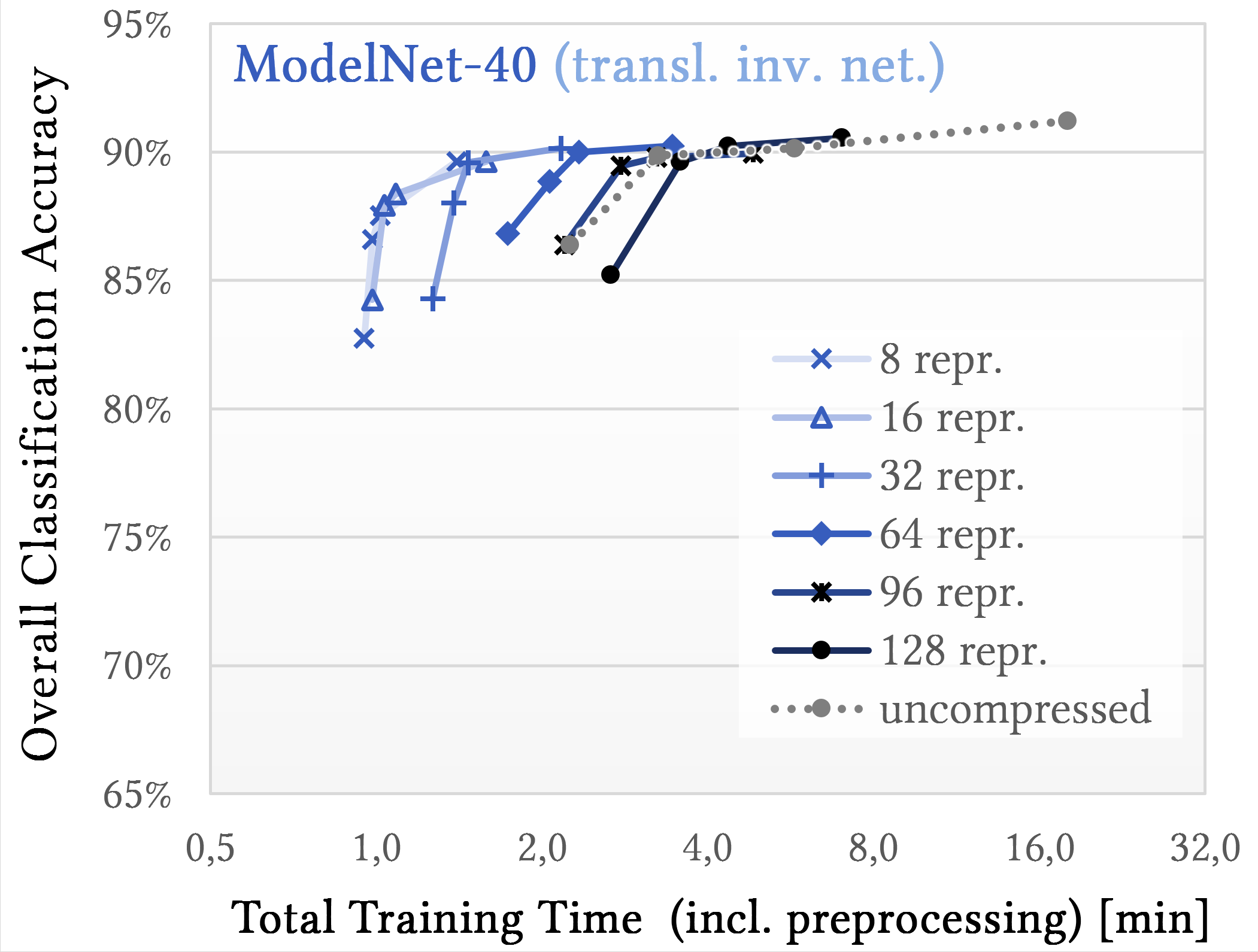}
    \caption{Performance comparison.}
    \label{fig:ModelNetTrPerformance}
  \end{subfigure}
  \caption{Evaluation of our translation-only equivariant classification architecture on the ModelNet40 test set. (a) shows the accuracy for different augmentation regimes for training and evaluation, while (b) and (c) use aligned inputs for both. (b) shows the accuracy loss due to compression vs the uncompressed network, and (c) shows the tradeoff between runtime and accuracy for different representative counts, as well as without compression.}
  \label{fig:ModelNetTr}
\end{figure}

\begin{figure}[H]
  \centering
  \begin{subfigure}[b]{0.33\linewidth}
    \includegraphics[width=\linewidth]{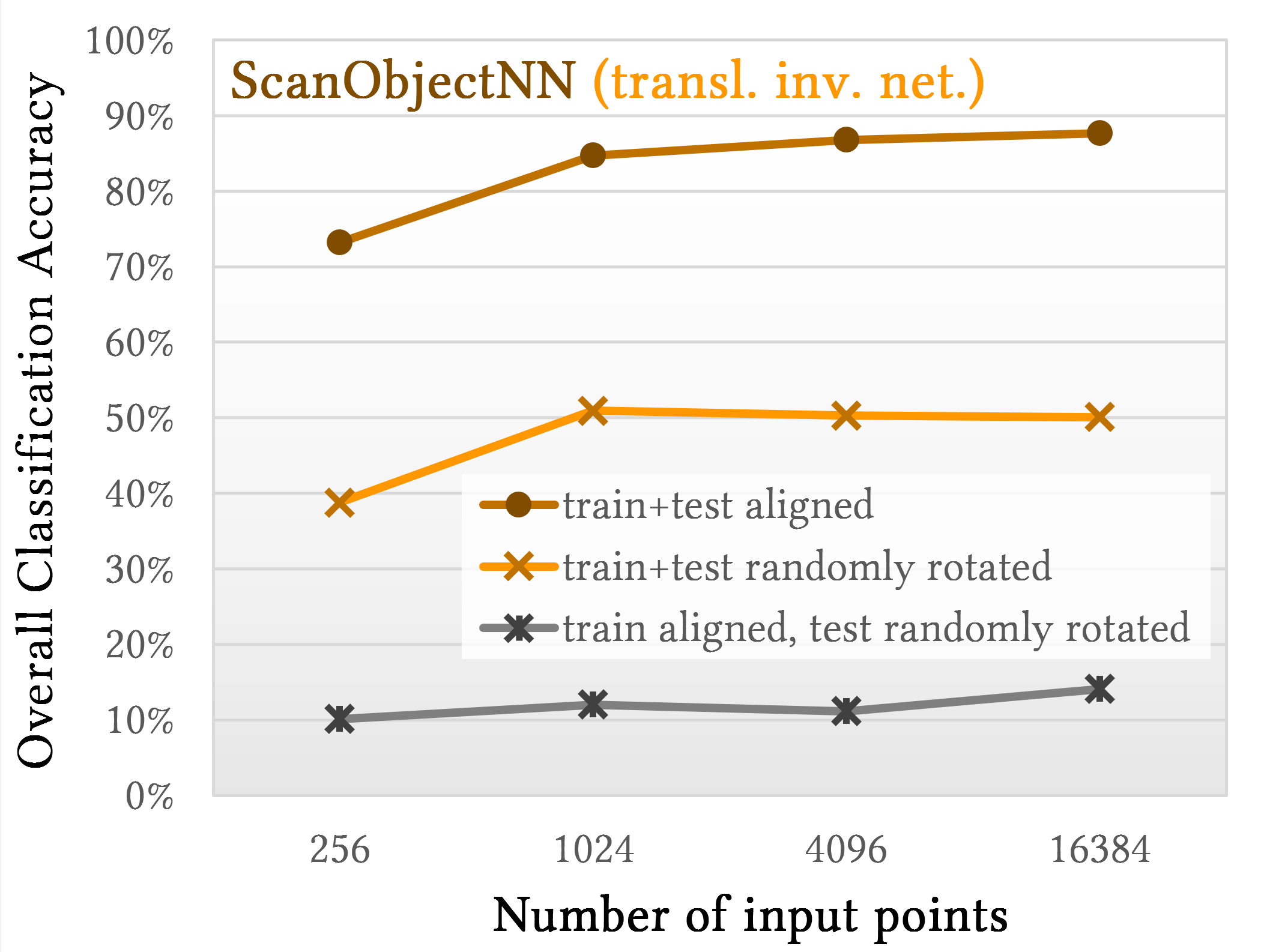}
    \caption{Uncompressed baseline.}
    \label{fig:ScanObjTrStability}
  \end{subfigure}
  \hfill
  \begin{subfigure}[b]{0.33\linewidth}
    \includegraphics[width=\linewidth]{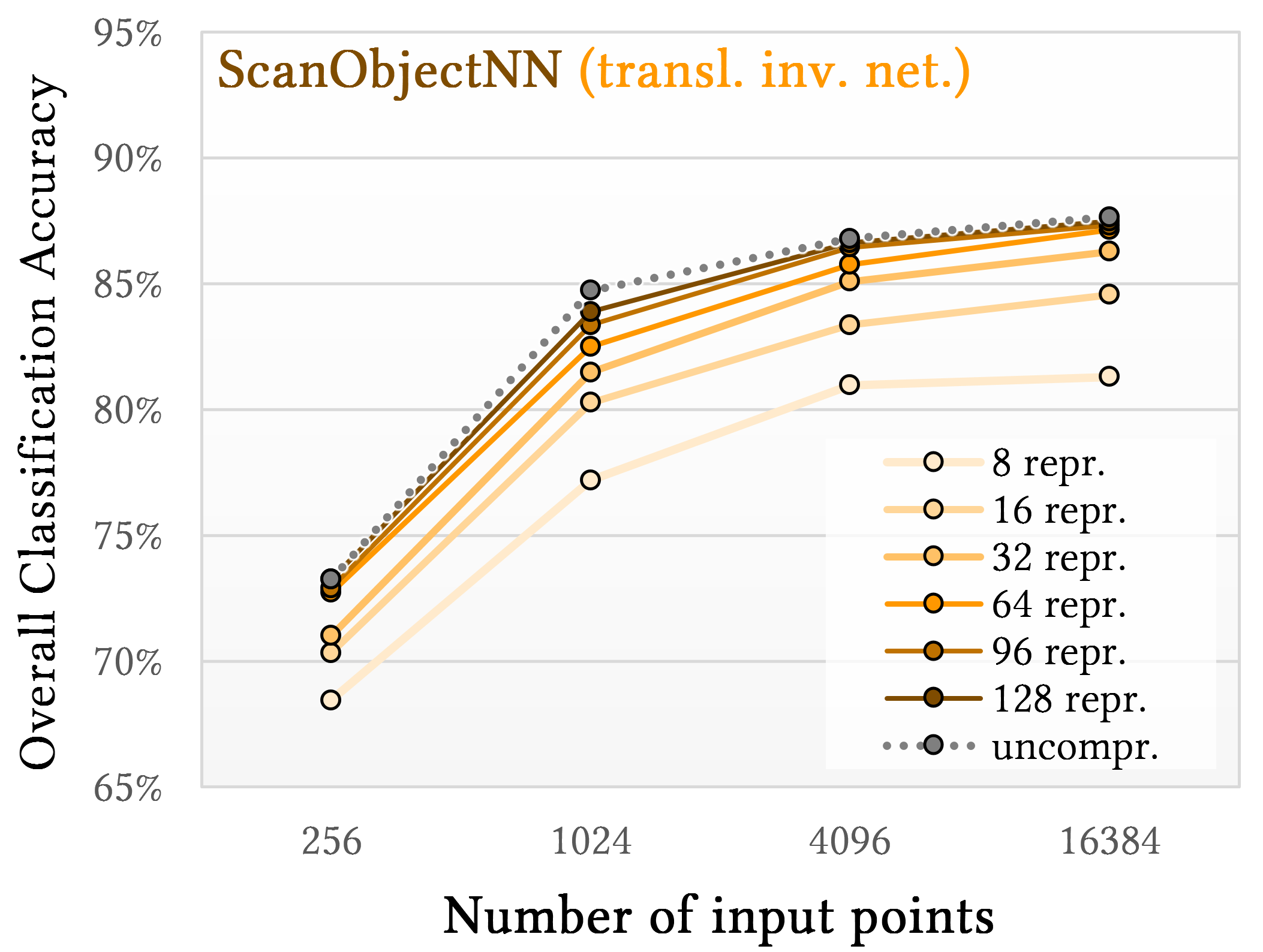}
    \caption{Compression accuracy loss.}
    \label{fig:ScanObjectTrAccuracy}
  \end{subfigure}
  \hfill
  \begin{subfigure}[b]{0.33\linewidth}
    \includegraphics[width=\linewidth]{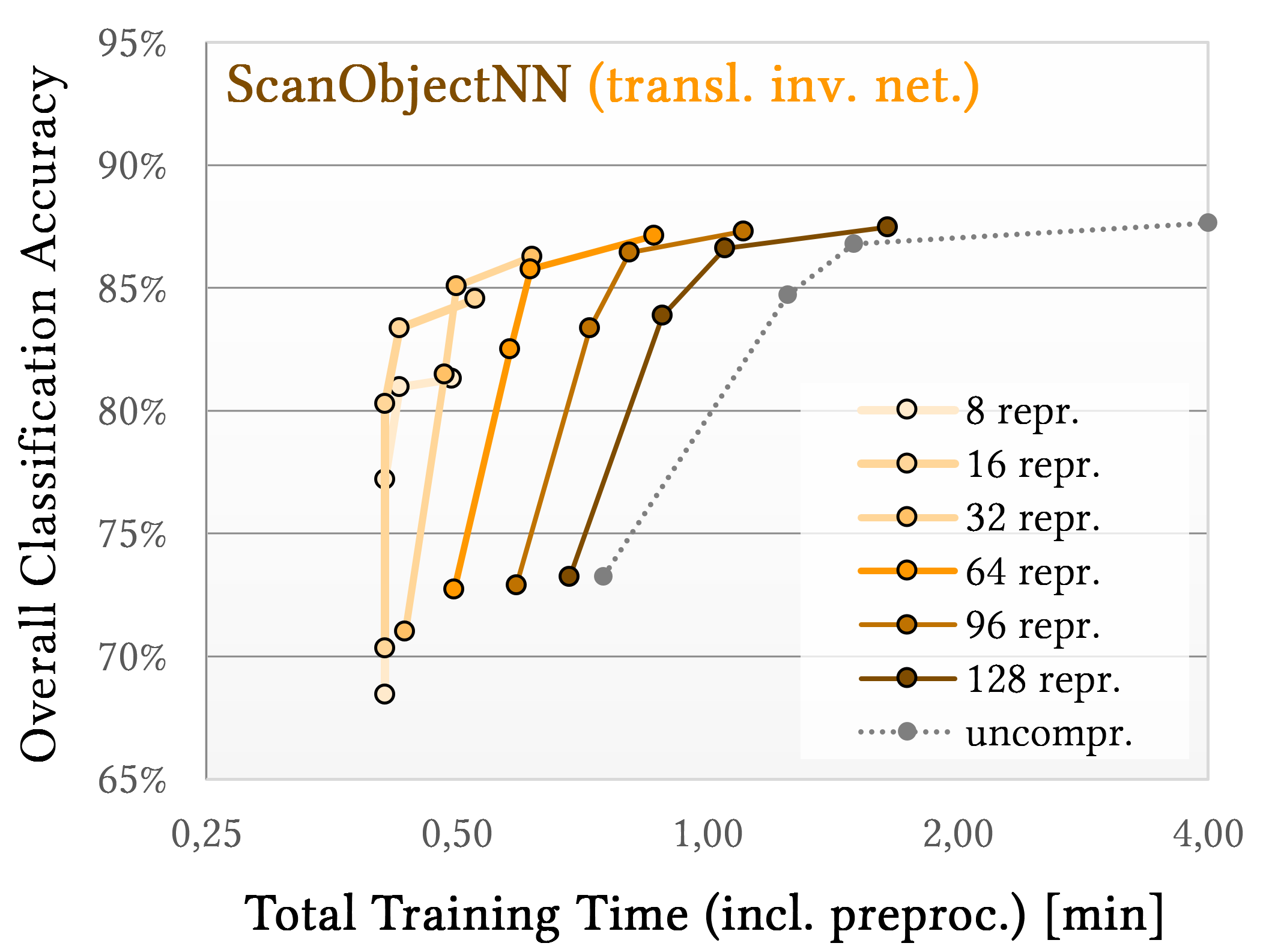}
    \caption{Performance comparison.}
    \label{fig:ScanObjectTrPerformance}
  \end{subfigure}
  \caption{Evaluation of our translation-only equivariant classification architecture on the ScanObjectNN test set. Same plots as in Fig.~\ref{fig:ModelNetTr}.}
  \label{fig:ScanObjTr}
\end{figure}

Fig.~\ref{fig:ModelNetTrAccuracy} and~\ref{fig:ScanObjectTrAccuracy} provide a detailed study of the impact of compression on accuracy. In Fig.~\ref{fig:ModelNetTrPerformance} and~\ref{fig:ScanObjectTrPerformance}, we plot accuracy against runtime for our compressed method across a range of representative counts.

Note that these results are obtained using aligned input point clouds during training and validation. Using randomly rotated inputs at validation causes a large performance loss with the translation-only equivariant network; even if the random rotations are also applied during training, accuracy will drop to $\sim 72\%$ for ModelNet (Fig.~\ref{fig:ModelNetTrStability}) and $\sim 50\%$ for ScanObjectNN (Fig.~\ref{fig:ScanObjTrStability}), highlighting the need for a rotationally equivariant network architecture.

\subsection{Classification with SE(3) equivariance}
\label{sec:results-classification-eq}
For full SE(3) equivariance, we test two different architectures: a full SO(3)-GCNN, and an optimized architecture using normal-aligned filters.

\paragraph{Full SO(3) GCNN:}
Here we use a set of 46 rotations randomly sampled by Poisson disc sampling on SO(3), similar to the approach of \citeN{kuipers2023se3}. For the construction of the full SO(3) GCNN architecture, we use these transformations as filter alignments (i.e., we use the Cartesian product of our (spatial) point cloud with all 46 rotations to form our new geometry representation), and we also form the Cartesian product of our $3 \times 3 \times 3$ spatial kernel with the 46 rotations to form our new SO(3) sampling kernel. This results in very large intermediate representations, so we can only run this method on our hardware using our compression technique. The results are shown in Table~\ref{tab:results} (iii). Even though we doubled the training epochs here, training accuracy is not saturating: for the run on ScanObject with 1024 spatial points, we get 68.0\% training accuracy and 63.8\% validation accuracy. This indicates that the network has difficulties learning in the fully equivariant regime. Still, it shows that our method allows us to actually perform the computation within the time and memory budget of our setup.

\paragraph{Optimized SE(3) equivariant model:} To construct a more efficient SE(3) equivariant model, we align our conv filters applied at each geometry point to surface normals of the geometry, following \citeN{Wiersma-SurfaceCNNs-Siggraph-2020} and \citeN{bekkers2024ponita}. This still leaves one degree of freedom for the filter alignment (rotation around the normal), which we sample using 8 equidistant rotations. We also switch to a systematic rotational sampling kernel by using the rotational octahedral group (i.e. the 24 rotations of a cube), following the construction of \citeN{worrall2018cubenet} and the more recent point-cloud variant of \citeN{zhu2023e2pn}. The optimized architecture works well on rotationally augmented data and achieves an accuracy close to that of the translation-only equivariant architecture on aligned data (see Table~\ref{tab:results}). However, training time is now much higher due to the larger point count and sampling kernel --- a single training epoch on ScanObjectNN now takes almost one hour. Our compression technique reduces the time per epoch to a range of 5 to 90 seconds, depending on the number of representatives. For a large enough representative count, i.e., 128, this has negligible impact on accuracy (85.59\% vs 85.42\%), and even when including the preprocessing step, our compression technique reduces the total training time by roughly a factor of ten compared to the uncompressed setting.

\begin{figure}[H]
  \centering
  \begin{subfigure}[b]{0.33\linewidth}
    \includegraphics[width=\linewidth]{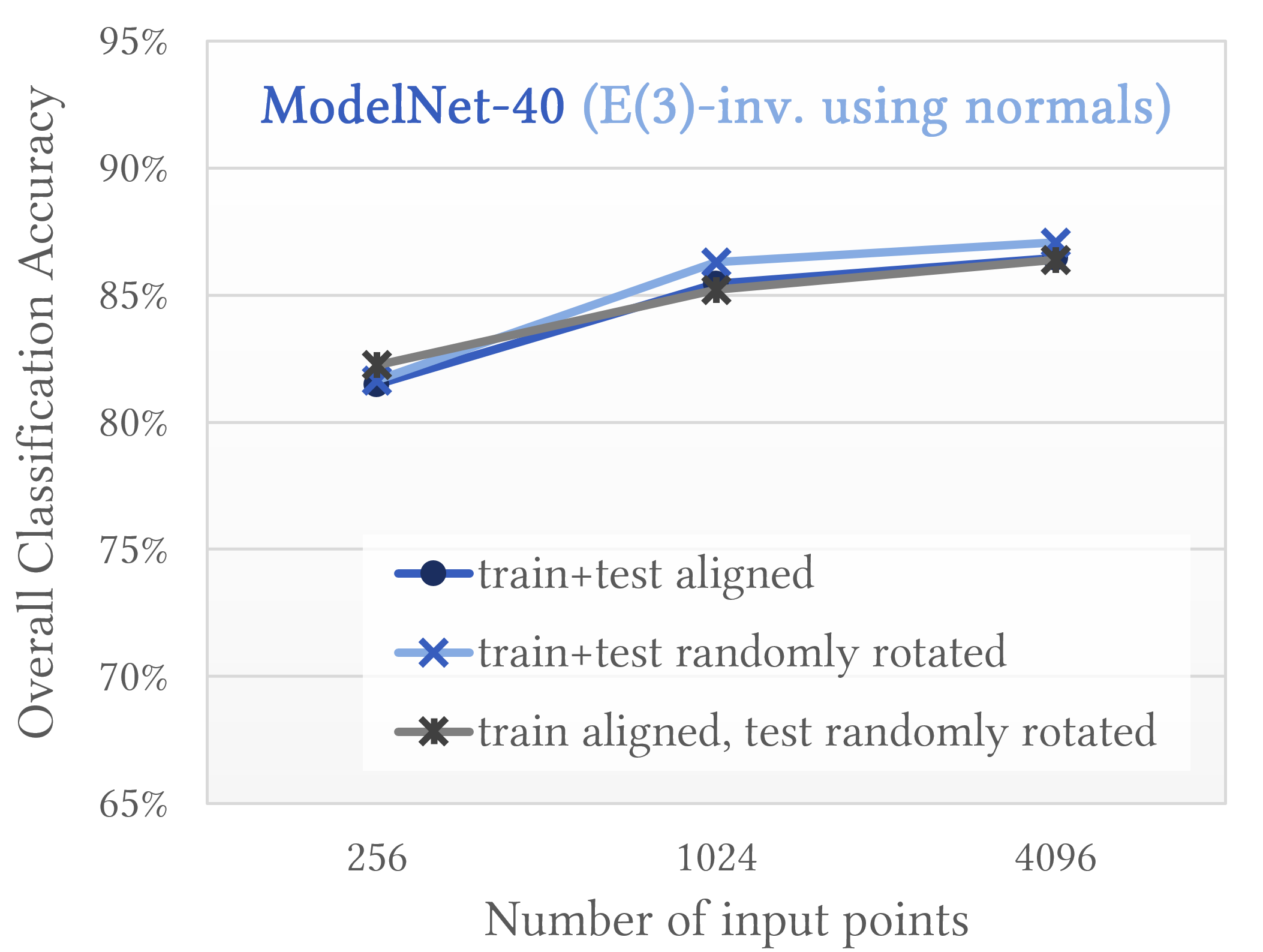}
    \caption{Uncompressed baseline.}
    \label{fig:ModelNetEqStability}
  \end{subfigure}
  \hfill
  \begin{subfigure}[b]{0.33\linewidth}
    \includegraphics[width=\linewidth]{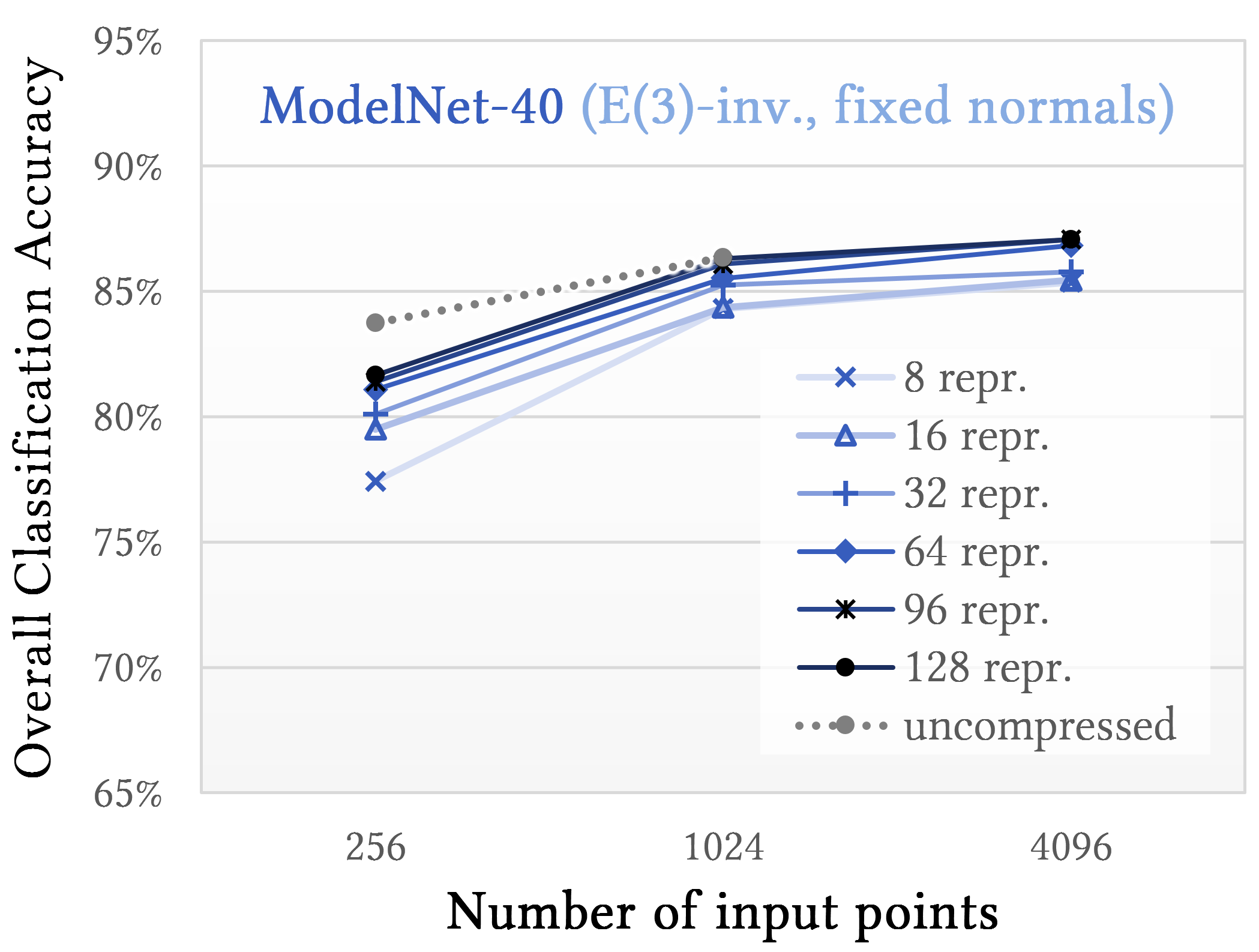}
    \caption{Compression accuracy loss.}
    \label{fig:ModelNetEqAccuracy}
  \end{subfigure}
  \hfill
  \begin{subfigure}[b]{0.33\linewidth}
    \includegraphics[width=\linewidth]{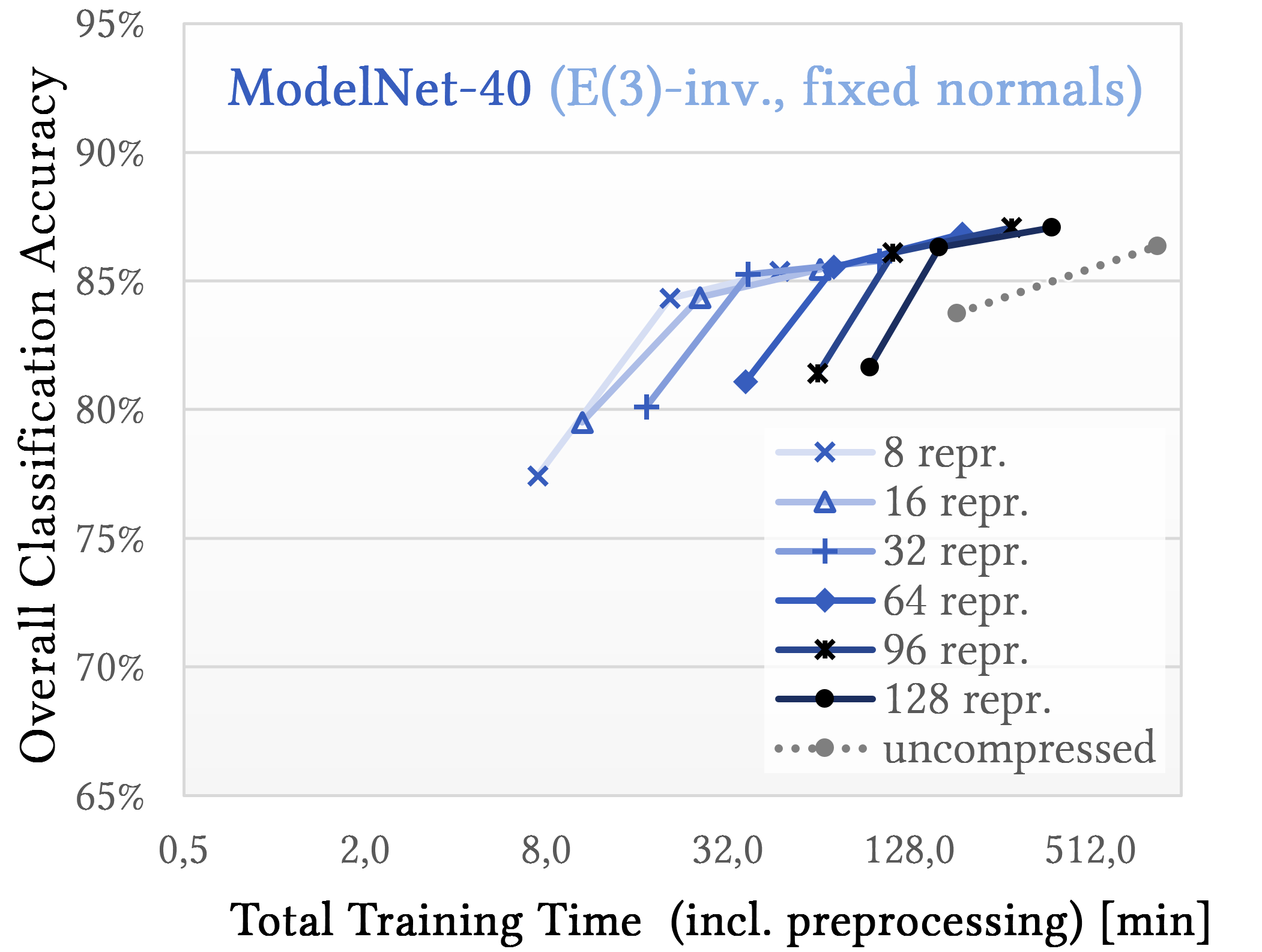}
    \caption{Performance comparison.}
    \label{fig:ModelNetEqPerformance}
  \end{subfigure}
  \caption{Evaluation of our surface-normal-aligned $SE(3)$-equivariant classification architecture on the ModelNet40 test set. (a) shows the accuracy for different augmentation regimes for training and evaluation, while (b) and (c) use randomly rotated inputs in both training and evaluation. (b) shows the accuracy loss due to compression vs the uncompressed network, and (c) shows the tradeoff between runtime and accuracy for different representative counts, as well as without compression. We omit the uncompressed training run at 4096 spatial points due to its long training time.}
  \label{fig:ModelNetEq}
\end{figure}

\begin{figure}[H]
  \centering
  \begin{subfigure}[b]{0.33\linewidth}
    \includegraphics[width=\linewidth]{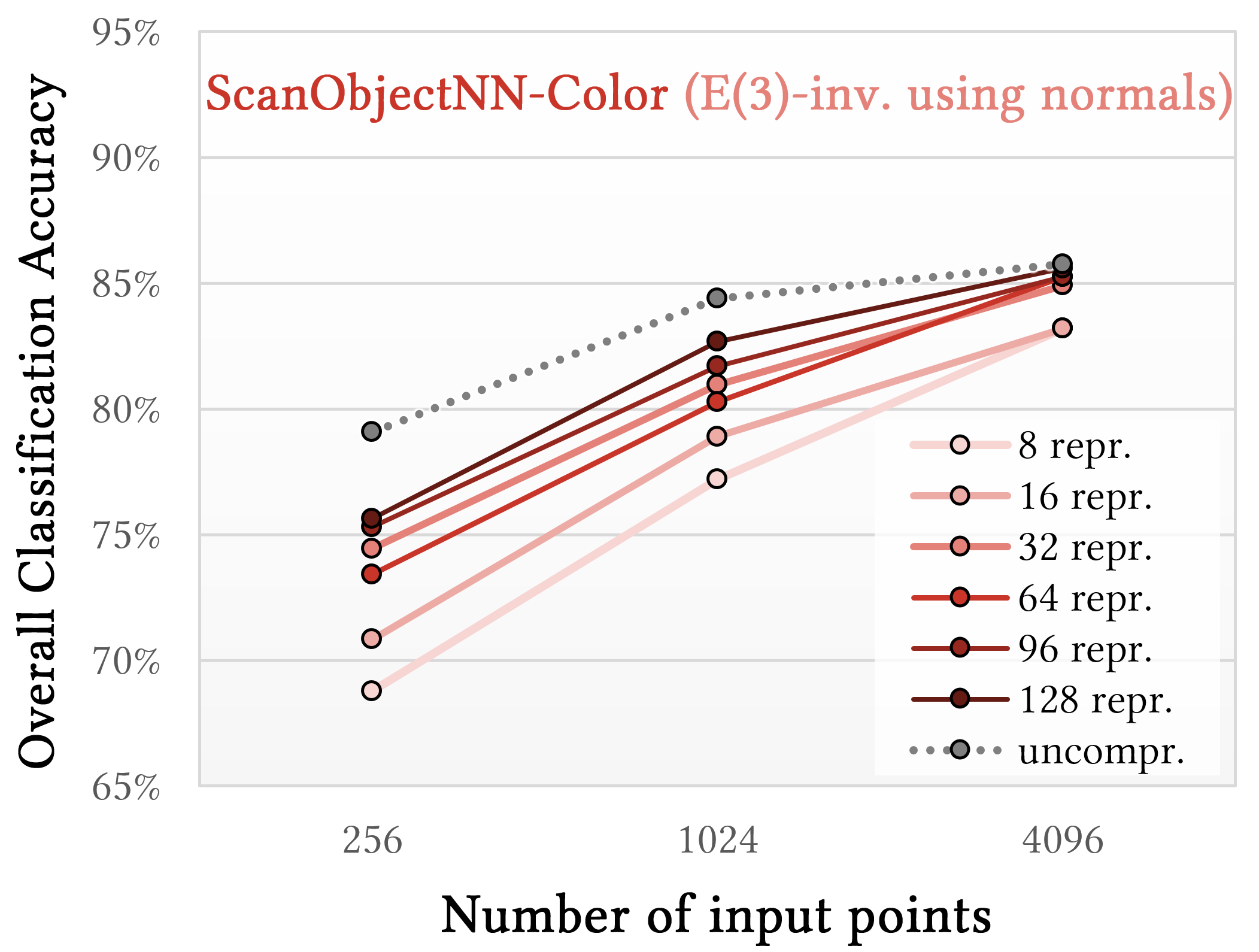}
    \caption{Compression accuracy loss (with colors).}
    \label{fig:ScanObjectEqAccuracyColor}
  \end{subfigure}
  \hfill
  \begin{subfigure}[b]{0.33\linewidth}
    \includegraphics[width=\linewidth]{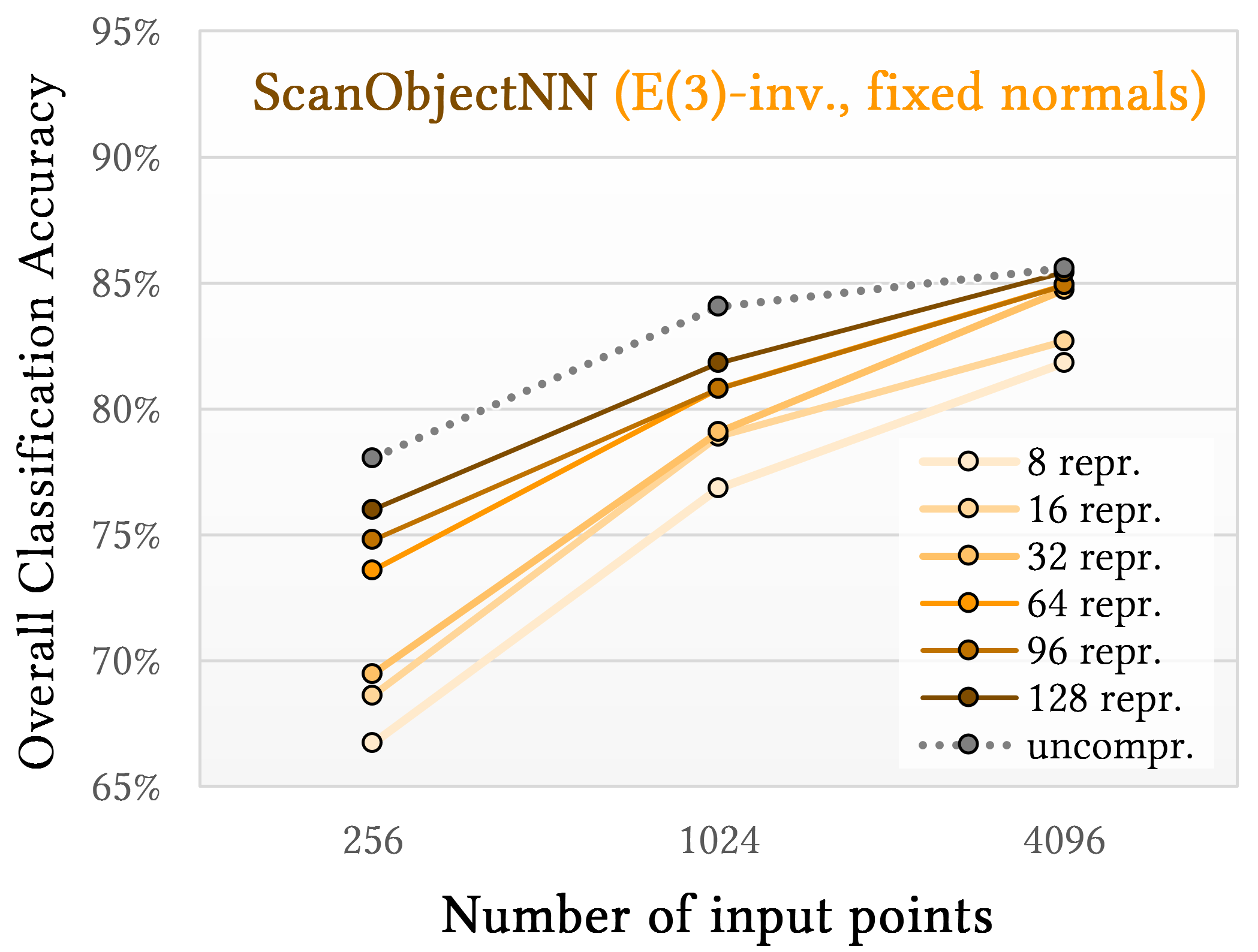}
    \caption{Compression accuracy loss (no colors).}
    \label{fig:ScanObjectEqAccuracy}
  \end{subfigure}
  \hfill
  \begin{subfigure}[b]{0.33\linewidth}
    \includegraphics[width=\linewidth]{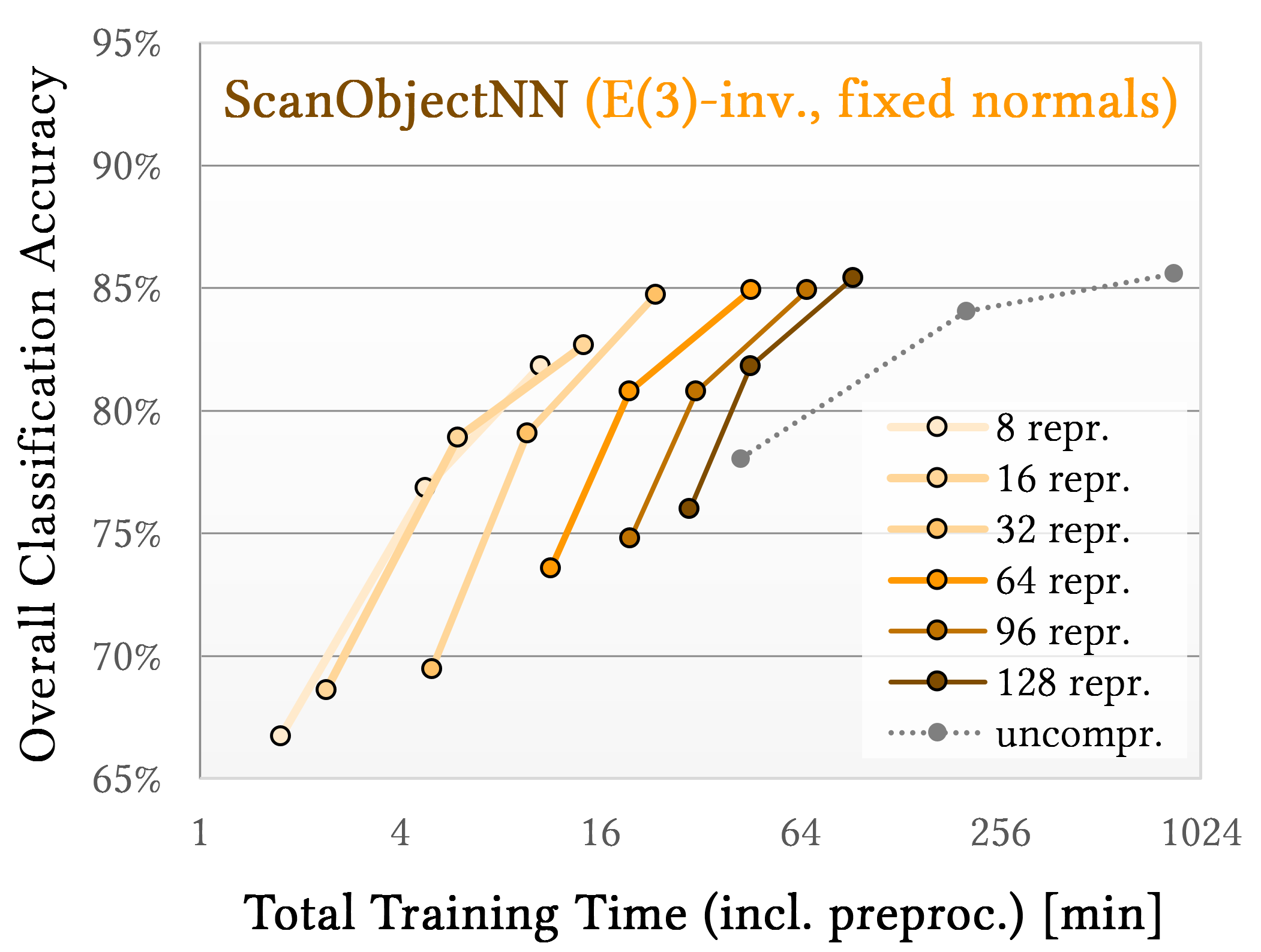}
    \caption{Performance comparison.}
    \label{fig:ScanObjectEqPerformance}
  \end{subfigure}
  \caption{Evaluation of our surface-normal-aligned $SE(3)$-equivariant classification architecture on the ScanObjectNN test set. The first two plots shows the accuracy loss due to compression vs the uncompressed network when using color information as input (a) and when using the pure geometry without color information (b). (c) shows the tradeoff between runtime and accuracy for different representative counts for the pure geometry case. All runs use randomly rotated inputs in both training and evaluation.}
  \label{fig:ScanObjEq}
\end{figure}

Even with a smaller number of representatives like 32 or 64 (see Fig.~\ref{fig:ModelNetEqAccuracy} and \ref{fig:ScanObjectEqAccuracy}), our compression technique allows us to achieve good accuracies. In turn, this permits an increase in the number of spatial sampling points. Even accounting for the preprocessing time, we outperform the baseline in terms of accuracy for any fixed total training time (see Fig.~\ref{fig:ModelNetEqPerformance} and~\ref{fig:ScanObjectEqPerformance}). Compression also works well when using the color information of ScanObjectNN dataset as additional input (Fig.~\ref{fig:ScanObjectEqAccuracyColor}), though color only seems to have a notable effect at low spatial sampling levels even before compression.

Our SE(3) equivariant architecture also provides good stability to rotation of the inputs: Even when trained on aligned data, accuracy on randomly rotated inputs does not drop substantially (Fig.~\ref{fig:ModelNetEqStability}).

\subsection{Ablation on Clustering}
\label{sec:clustering-ablation}

\label{sec:where-to-cluster}
\begin{figure}[H]
  \centering
  \begin{subfigure}[b]{0.33\linewidth}
    \includegraphics[width=\linewidth]{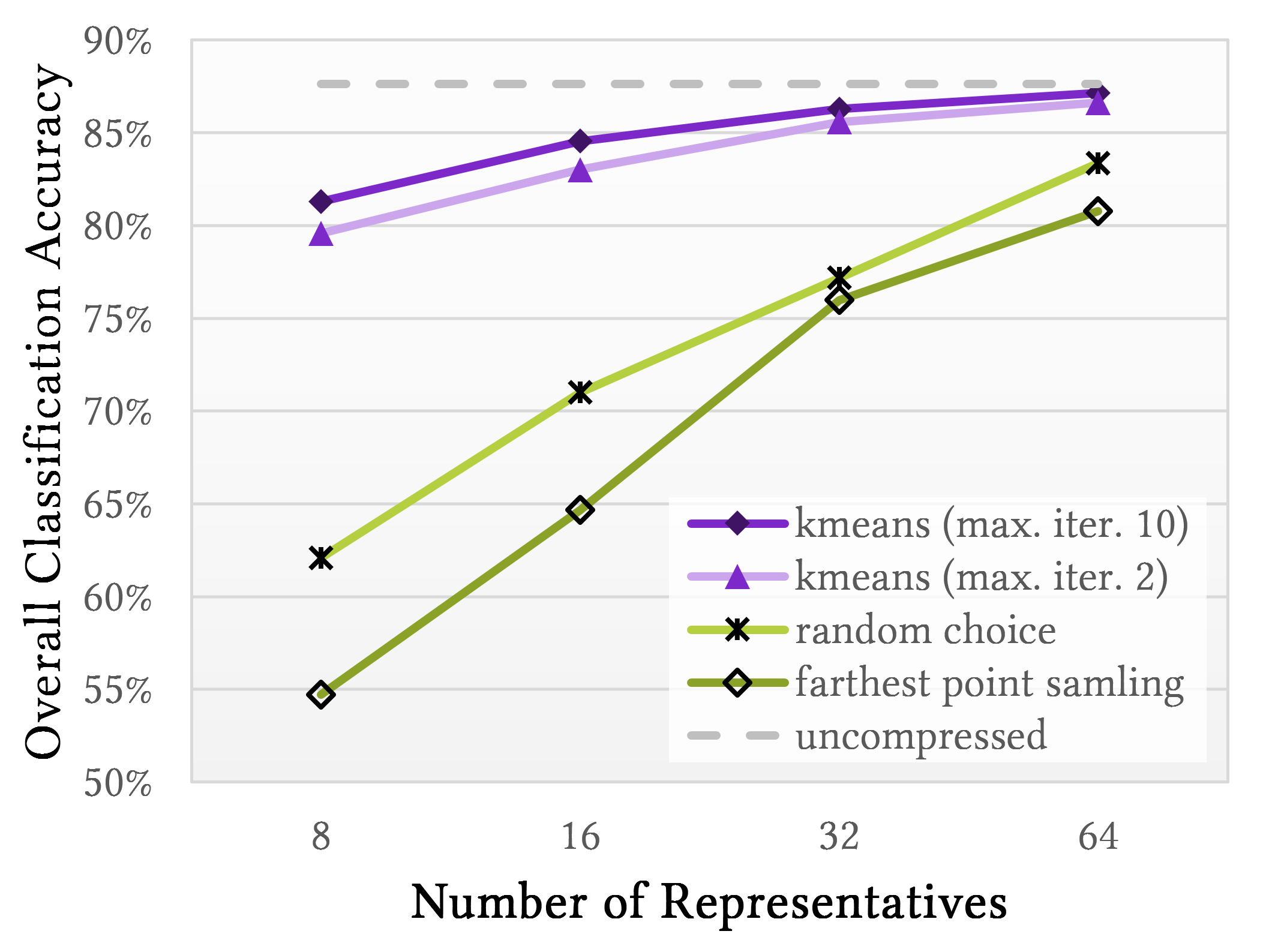}
    \caption{Different clustering algorithms.}
    \label{fig:ScanObjAblationAlgo}
  \end{subfigure}
  \hfill
  \begin{subfigure}[b]{0.33\linewidth}
    \includegraphics[width=\linewidth]{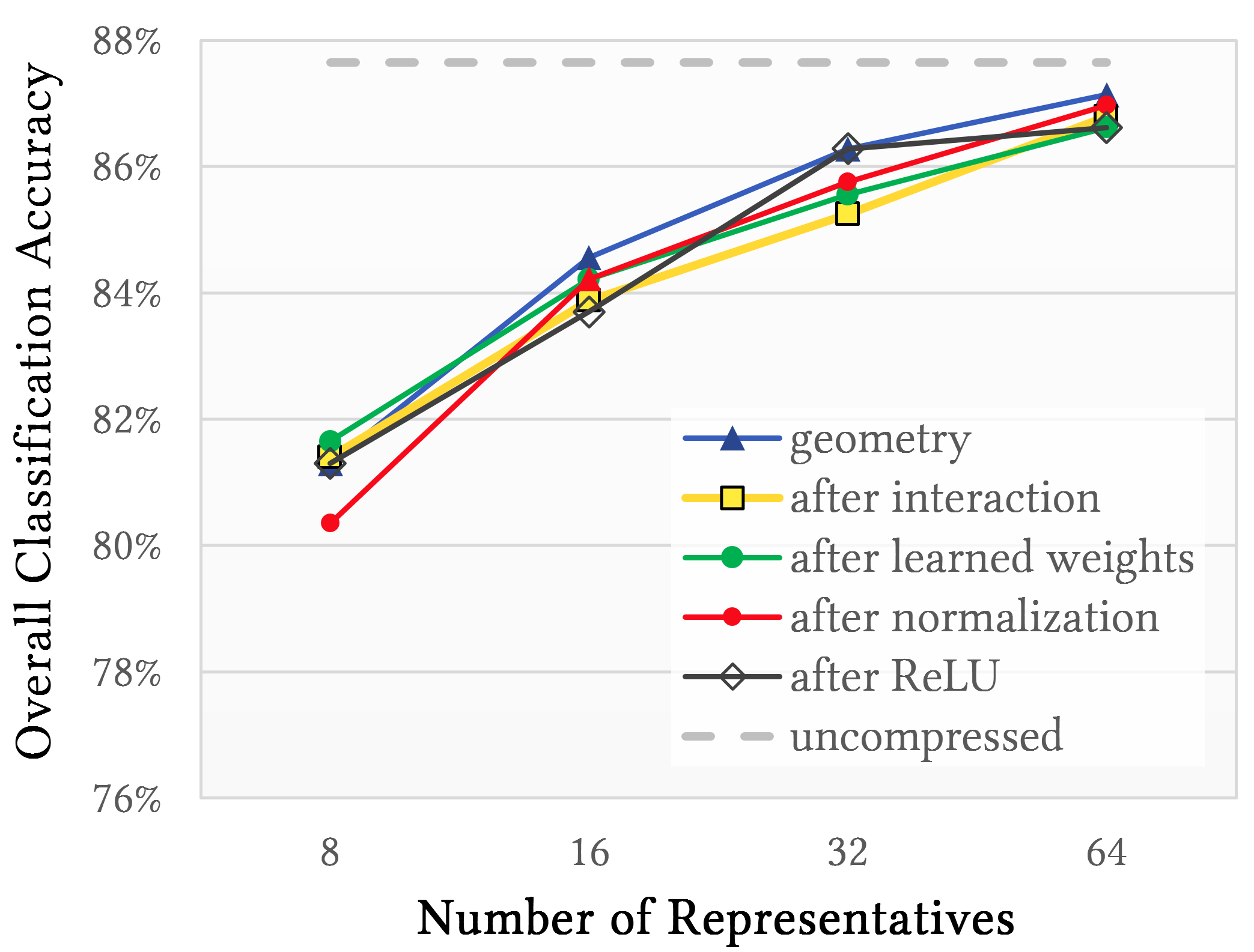}
    \caption{Different targets, clustered in first epoch only.}
    \label{fig:ScanObjAblationNoRecluster}
  \end{subfigure}
  \hfill
  \begin{subfigure}[b]{0.33\linewidth}
    \includegraphics[width=\linewidth]{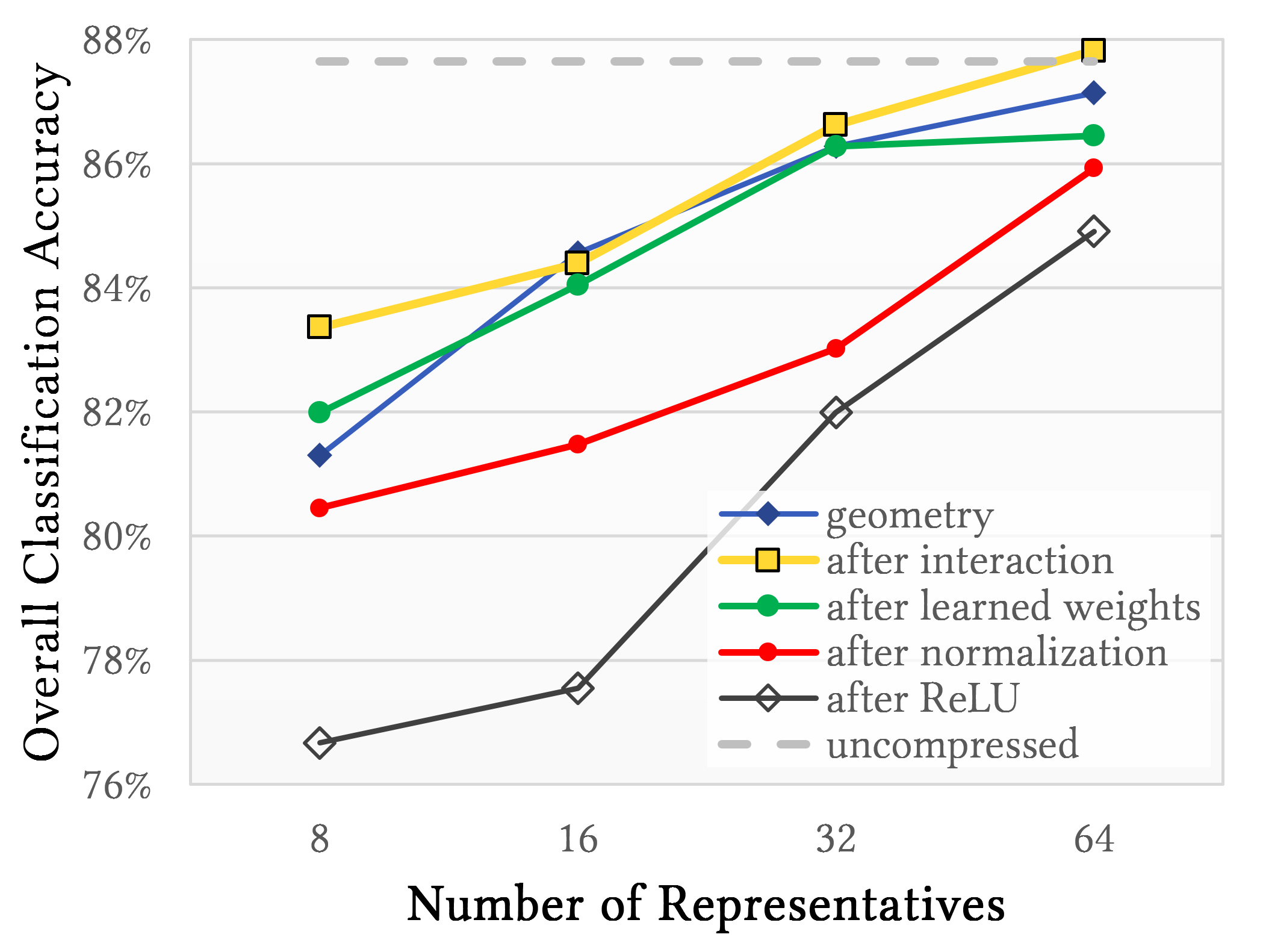}
    \caption{Different targets, reclustered in each epoch.}
    \label{fig:ScanObjAblationRecluster}
  \end{subfigure}
  \caption{
  Comparison of clustering strategies on ScanObjectNN (16384 points, 64 representatives, translation-only equivariant model). The ablation study shows in particular that geometry-based preprocessing is competitive with reclustering in each epoch. For a detailed description of the experiment, see Table~\ref{tab:clustering}.}
  \label{fig:ScanObjAblation}
\end{figure}

\paragraph{Where to Cluster.}
We compare the five clustering targets defined in Section~\ref{sec:clustering}: clustering on the post-interaction activations $h_{n,k,c}$, on the post-weight activations
$\hat\phi'_{n,c'}$, on post-batch-normalization activations, on post-ReLU activations, and the geometry-only scheme based on one-hot cluster ids from the previous layer.
Table~\ref{tab:clustering} and Fig.~\ref{fig:ScanObjAblationNoRecluster} report classification accuracy for each variant. The four activation-based options yield very similar accuracies, and the geo\-metry-only scheme matches or slightly exceeds them while remaining independent of any trained weights. We therefore chose geometry-only as our default.

\label{sec:clustering-algorithm}
\paragraph{Choice of Clustering Algorithm.}
We compare several clustering algorithms with respect to both accuracy and preprocessing cost; an overview is given in Table~\ref{tab:clustering} and  Figure~\ref{fig:ScanObjAblationAlgo}. We find that $k$-means with a small number of iterations offers the best trade-off. Already at a maximum of two iterations, classification accuracy is close to that of fully converged $k$-means; a few additional iterations yield a small further improvement, and we settle on a maximum of ten iterations as our default. Choosing cluster centers as a random subset of the points removes nearly all clustering overhead—it requires only a single nearest-representative assignment—but yields markedly worse accuracy. It can be viewed as a special case of $k$-means with \texttt{max\_iter}$=0$, since $k$-means likewise initializes its centers randomly. Farthest-point sampling performs worse still and is by a substantial margin the slowest of the methods we evaluate.

\paragraph{Reclustering.}
Cluster assignments can in principle be recomputed each epoch. Table~\ref{tab:clustering} (bottom, ``Each ep.'' column) shows this helps slightly when clustering on $h_{n,k,c}$ (86.79\% $\rightarrow$ 87.82\%), but hurts in all other cases, with the loss growing the later clustering happens. We include reclustering only as an analytical baseline; it eliminates the runtime advantage entirely.

\begin{table}[t]
  \centering
  \caption{Clustering ablation on ScanObjectNN (16384 points, 64 representatives, translation-only equivariant model). Top: different clustering algorithms applied to geometry-only features. Bottom: different points in the convolution block at which to cluster. The ``1st ep.'' column reports accuracy when assignments are computed once in the first training epoch and reused; the ``Each ep.'' column reports accuracy when assignments are recomputed every epoch. Runtime is reported for a full training epoch including clustering. See Fig.~\ref{fig:ScanObjAblation} for the effect of varying the representative count.}
  \label{tab:clustering}
  \small
  \setlength{\tabcolsep}{3pt}
  \begin{tabular}{l l c c c}
    \toprule
    Cluster by & Algorithm & 1st ep. & Each ep. & Time (s) \\
    \midrule
    geometry & $k$-means (10 iter.)       & 87.14\% & ---     & 30 \\
    geometry & $k$-means (2 iter.)        & 86.62\% & ---     & 25 \\
    geometry & random choice              & 83.36\% & ---     & 23 \\
    geometry & farthest point sampling    & 80.79\% & ---     & 90 \\
    \midrule
    after interaction     & $k$-means (10 iter.) & 86.79\% & 87.82\% & 28 \\
    after weights         & $k$-means (10 iter.) & 86.62\% & 86.45\% & 25 \\
    after normalization   & $k$-means (10 iter.) & 86.97\% & 85.93\% & 25 \\
    after nonlinearity    & $k$-means (10 iter.) & 86.62\% & 84.91\% & 25 \\
    \bottomrule
  \end{tabular}
\end{table}

\subsection{Segmentation experiments}
\label{sec:segmentation-experiments}
\begin{figure}[H]
  \centering
  \begin{subfigure}[b]{0.33\linewidth}
    \includegraphics[width=\linewidth]{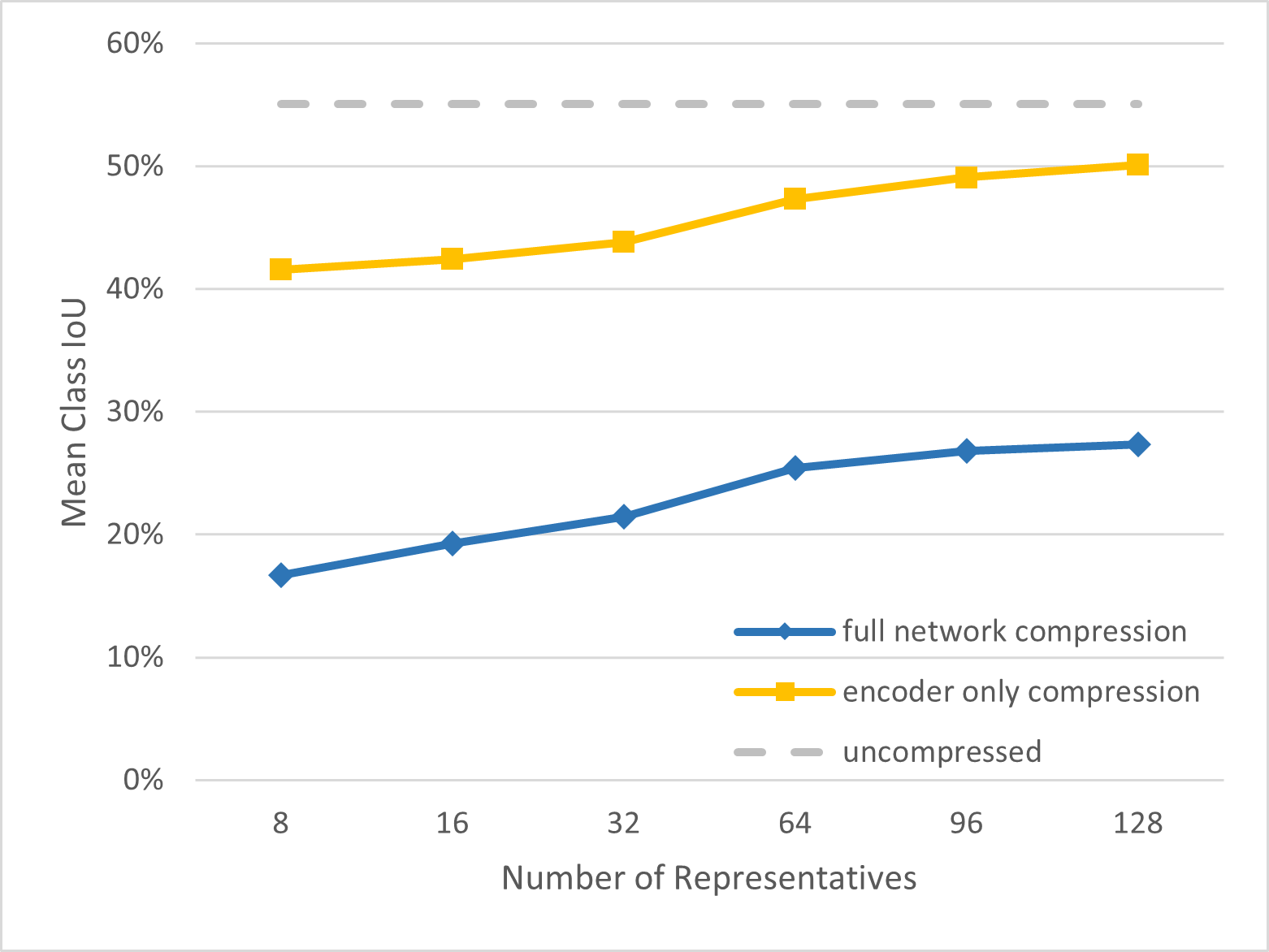}
    \caption{Uncompressed vs compressed performance.}
    \label{fig:UnetAblationEncDec}
  \end{subfigure}
  \hfill
  \begin{subfigure}[b]{0.33\linewidth}
    \includegraphics[width=\linewidth]{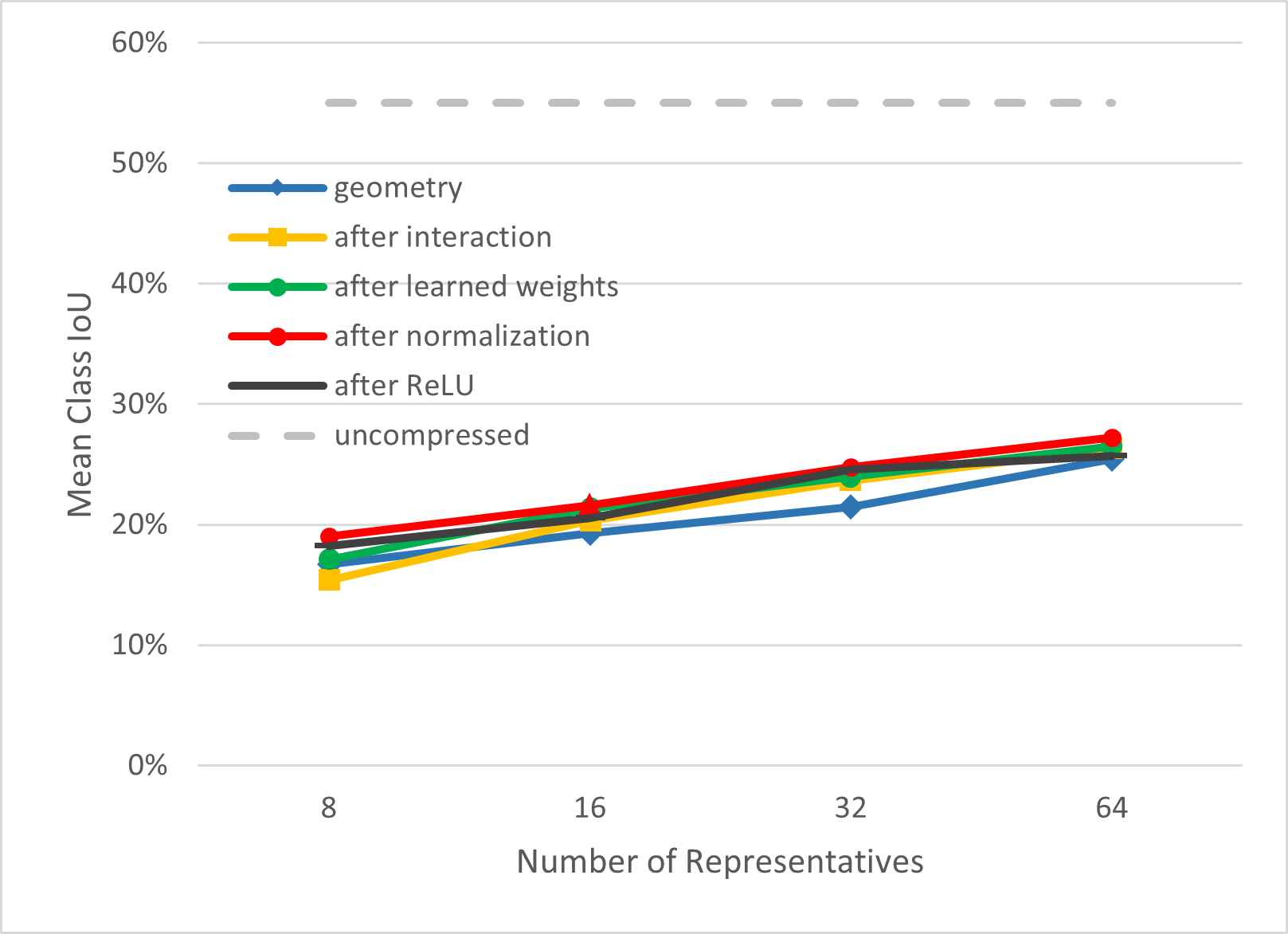}
    \caption{Different targets, reclustered in each epoch.}
    \label{fig:UnetAblationNoRecluster}
  \end{subfigure}
  \hfill
  \begin{subfigure}[b]{0.33\linewidth}
    \includegraphics[width=\linewidth]{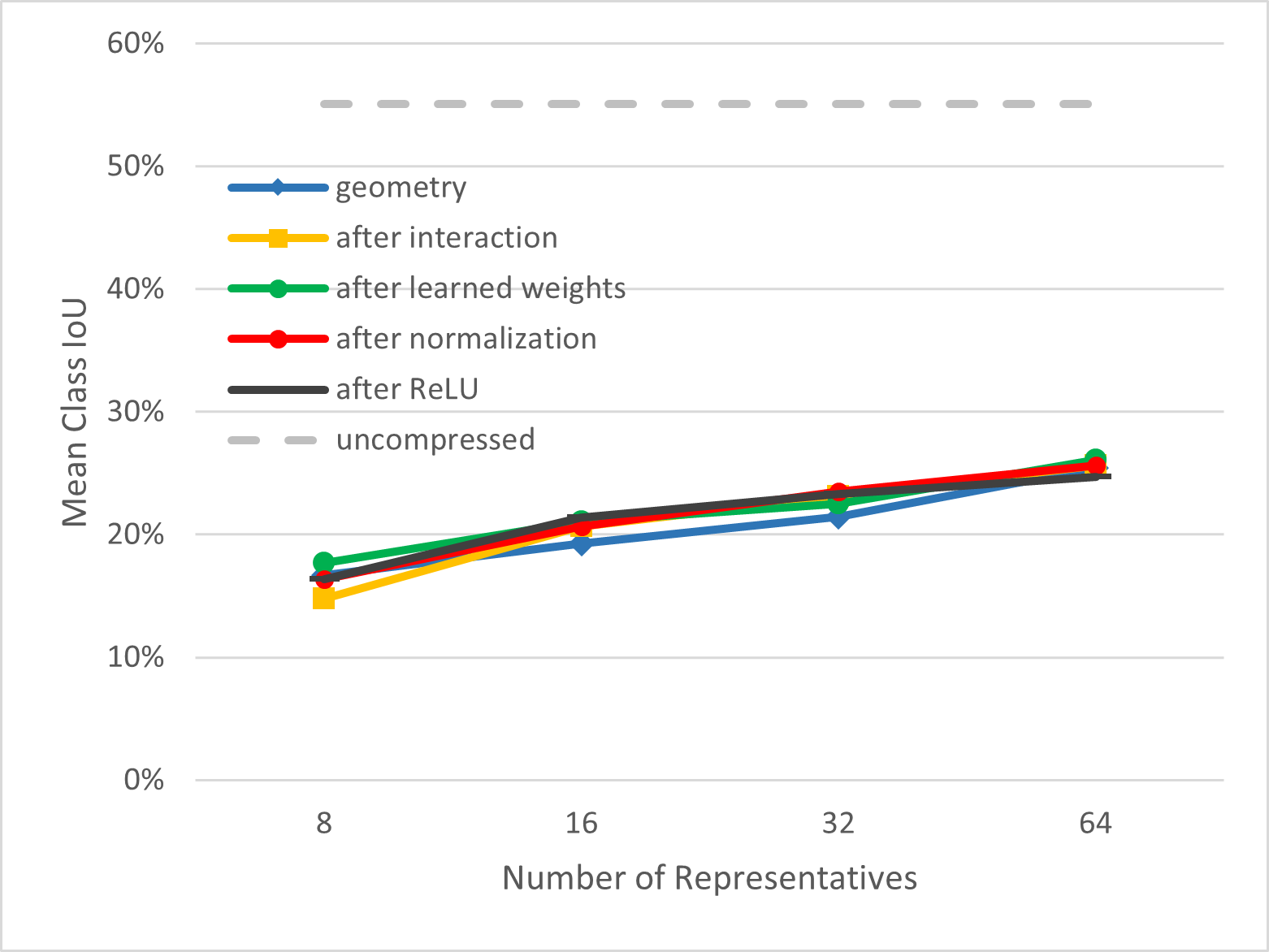}
    \caption{Performance for encoder-only compression.}
    \label{fig:UnetAblationRecluster}
  \end{subfigure}
  \caption{Results for our segmentation experiment on the ScanNet20 validation set (not using color information). Note that the plots report mean class IoU instead of accuracy. (a) compares the uncompressed network vs (partial) compressed variant. (b) and (c) investigate the effect of different clustering targets for the fully compressed network. Dynamic reclustering in each epoch does not improve results here, either.}
  \label{fig:UnetAblation}
\end{figure}

Our experiments with the U-Net architecture on ScanNet (Fig.~\ref{fig:UnetAblationEncDec}) show our compression method to work well when only used in the encoder path of the network (mIoU of 0.50 vs 0.55 uncompressed), but applying compression to the full network (encoder and decoder path) leads to a large performance drop (mIoU of 0.27). We attribute this gap to the structure of decoder features, which combine upsampled deep features with encoder skip connections and consequently exhibit a much broader distribution that does not compress as well. We found that switching the clustering criterion from geometry to activations (Fig.~\ref{fig:UnetAblationNoRecluster}) leads to a slight improvement for the fully compressed network, but does not close the gap in performance.

\section{Conclusions}

We presented a compression method for group-convolutional neural networks based on linear representation theory. In traditional implementations, costs scale exponentially in the degrees of freedom of the symmetry group (dimension of the Lie-algebra). Even for moderate scenarios, such as including a single rotational degree of freedom to a surface-aligned network, memory and runtime costs become problematic, and fully sampling three rotational degrees of freedom exceeds the capabilities of a high-end consumer system on ModelNet or ScanObjectNN. Our paper explores the idea of discretizing networks in feature space rather than spatially, effectively compressing samples of similar geometry into single representatives. While gains are moderate for purely translational CNNs, rotational equivariance puts the compressed method ahead of a traditional implementation in training time with and without preprocessing, and, by permitting finer resolutions, can also enable higher accuracy figures.
Overall, we believe that our paper shows a useful alternative method for discretizing linear group representations for deep learning, opening a new direction for efficient equivariant networks that complements existing efficiency techniques. %

\subsection{Limitations and Future Work}
While decoupling costs from degrees of freedom is promising, our approach has several limitations. First, effectiveness strongly depends on the application. While CNN-classifiers with incremental downsampling yield strong savings, clustering does not work well for the segmentation task, as seen in Section~\ref{sec:segmentation-experiments}: The U-Net architecture, specifically, has a high-resolution output field where every output surfel has a large receptive field, which reduces compression efficiency. We expect the broader limitation---poor compressibility of features that mix information across spatial scales---to apply more generally to architectures with similar feature flow, likely including transformer-based point cloud models with global attention, although we have not evaluated this explicitly. For such architectures, a more complex compression scheme that takes activation statistics during training into account might be a path forward.

We also restrict our study to a very simple CNN-network that does not reach state-of-the-art accuracies. While our setup clearly isolates the key effects, a more involved setting might reveal additional trade-offs. This is particularly the case for transformer-type networks, which we do not examine in this paper. Arguably, any positional encoding based on linear group representations might also lend itself to similar types of treatment; a detailed assessment remains subject to future work. Finally, our preprocessing method for finding partial symmetry is scalable but remains basic; it would be great to explore processing of very large scenes (such as city-scale LIDAR scans) using a highly-optimized approach.

\onecolumn\clearpage

\vfill

\vfill

\clearpage

\vfill

\vfill

\twocolumn\clearpage

\section*{Acknowledgment}
\paragraph{Funding}
This work was supported by the Deutsche Forschungsgemeinschaft (DFG, German Research Foundation), project 233630050 (Collaborative Research Center TRR146).

\paragraph {AI tool usage}
Generative AI language tools were used for copy-editing assistance, including grammar corrections, spelling, and rephrasing suggestions. No new content was generated by AI; all output was reviewed, revised, and validated by the authors, who assume full responsibility for this manuscript.

\bibliographystyle{ACM-Reference-Format}
\bibliography{references}

\clearpage

\end{document}